\ifdefined\XeTeXversion\else\ifdefined\directlua\else\pdfoutput=1\fi\fi 
\documentclass[11pt]{article}
\usepackage{thefinai}

\usepackage{pifont}       
\usepackage{textalpha}   
\newcommand{\greekfont}{}
\usetikzlibrary{arrows.meta,positioning}
\newcommand{\cmark}{\textcolor{green!60!black}{\ding{51}}} 
\newcommand{\xmark}{\textcolor{red}{\ding{55}}}            
\newcommand{\bench}{\textsc{MultiFinGround}}
\newcommand{\concept}[1]{\texttt{#1}}
\newcommand{\pending}{\textendash}

\newcommand{\target}[1]{#1}

\title{Recognition Is Not Grounding: Benchmarking Multilingual Financial Evidence Extraction}
\date{\today}

\author{%
  Lingfei Qian\textsuperscript{1}, Yan Wang\textsuperscript{1,*}, Xueqing Peng\textsuperscript{1,*},
  Polydoros Giannouris\textsuperscript{2}, Nadhem ZMANDAR\textsuperscript{1}, Yi Han\textsuperscript{3},
  Yanguang Liu\textsuperscript{4}, Sophia Ananiadou\textsuperscript{2}, Víctor Gutiérrez-Basulto\textsuperscript{5}%
}
\affiliation[1]{The Fin AI, USA}
\affiliation[2]{University of Manchester, UK}
\affiliation[3]{Georgia Institute of Technology, USA}
\affiliation[4]{New Jersey Institute of Technology, USA}
\affiliation[5]{Cardiff University, UK}
\addlogo[2.6em]{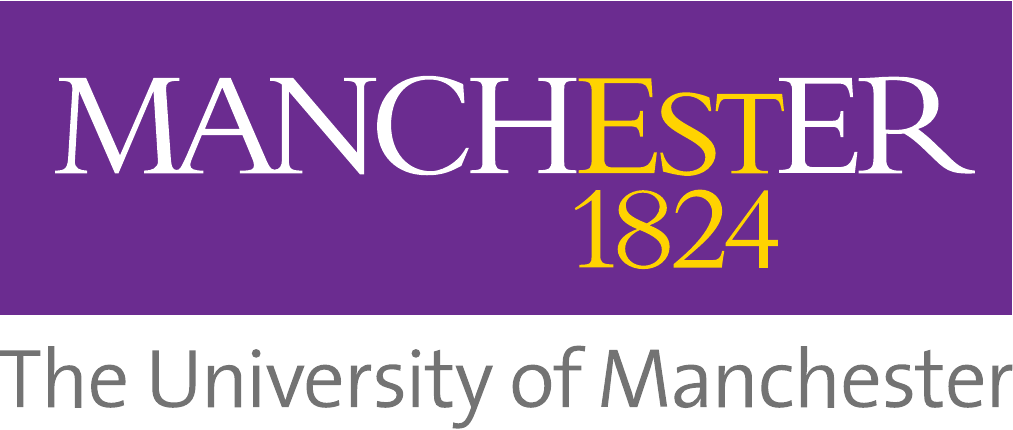}
\addlogo[2.4em]{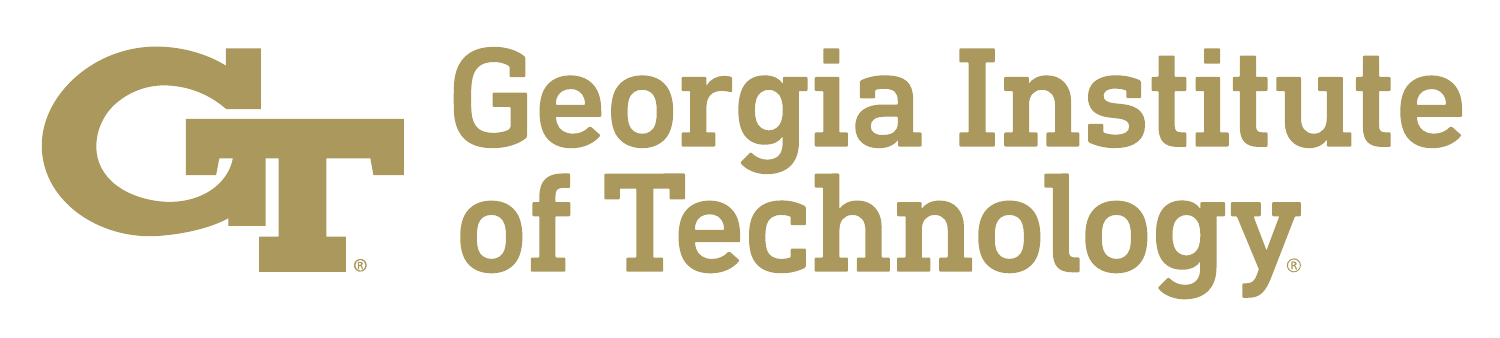}
\addlogo[2.6em]{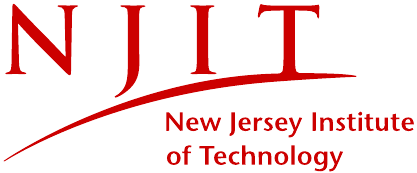}
\addlogo[2.6em]{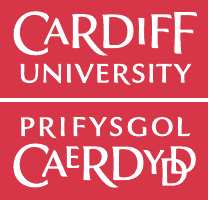}
\authornote{\textsuperscript{*}Corresponding authors}

\correspondence{\href{mailto:wy2266336@gmail.com}{wy2266336@gmail.com}, \href{mailto:xueqing.peng2024@gmail.com}{xueqing.peng2024@gmail.com}}

\begin{document}

\begin{abstract}
Comparing financial disclosures across languages requires identifying shared accounting concepts and recovering all numerical facts that express them. Existing benchmarks evaluate aspects of these capabilities, but provide limited insight into whether both succeed on the same multilingual evidence. In particular, it remains unclear whether language models can recognize these concepts across languages and exhaustively recover their corresponding numerical evidence from the same contexts.
We introduce \bench{}, a benchmark designed to investigate this recognition–grounding gap in multilingual financial reports. Constructed from FY2025 disclosures of 90 issuers in English, French, and Greek, \bench{} aligns independently authored reports through a shared accounting taxonomy and provides traceable fact-level annotations. The benchmark evaluates financial fact identification, concept recognition, exhaustive fact selection, and concept discrimination without candidate labels. 
Across 15 evaluated models, eight API and open-weight reference models achieve 97.3–99.4\% concept recognition accuracy, yet miss at least one target fact in 26.6–41.9\% of correctly recognized groups. Further analyses reveal substantial variation in concept discrimination and limited gains from the evaluated finance-specialized models. Our findings demonstrate that strong concept recognition can coexist with incomplete evidence recovery, highlighting the importance of evaluating semantic recognition and evidence completeness together.
\end{abstract}

\maketitle

\section{Introduction}

Investors and analysts comparing companies across countries need comparable financial information from reports published in different languages \citep{esmaElectronic}. Multilingual financial information extraction therefore requires models to find numerical facts that express shared accounting concepts across languages and reporting formats. However, these values can appear under different terms, in different table structures, or for different reporting periods  \citep{zhao2022multihiertt}. Consider a French report giving revenue of EUR 280 million for the current year and EUR 250 million for the previous year. A model may correctly recognize the revenue concept yet retrieve only the current-year amount, omitting the previous-year value needed to calculate annual growth. The current-year figure is correct, but the missing comparison prevents calculation of annual growth. This illustrates a challenge in multilingual numerical grounding: recognizing a concept does not guarantee complete recovery of its numerical evidence \citep{reddy2024docfinqa,choe2025hierarchical}.

Existing benchmarks assess concept linking, extraction, and reasoning (Table~\ref{tab:benchmark-comparison}). These works like FiNER, FNXL, and FinTagging evaluate financial numeral tagging and taxonomy linking \citep{loukas-etal-2022-finer,sharma-etal-2023-financial,wang2026fintagging}; FinQA and TAT-QA test numerical reasoning \citep{chen-etal-2021-finqa,zhu-etal-2021-tat}. MultiFin evaluates multilingual financial classification \citep{jorgensen-etal-2023-multifin}, while MultiFinBen and XTREME-R include multilingual extraction tasks \citep{peng-etal-2026-multifinben,ruder-etal-2021-xtreme}. 
However, these works provide limited insight into whether models can recognize shared accounting concepts across languages, recover all corresponding numerical facts, and how performance on these two tasks relates. Examining these questions requires paired evaluation of concept recognition and exhaustive fact selection on the same multilingual contexts.

\begin{table*}[t]
\centering\small
\setlength{\tabcolsep}{4pt}
\renewcommand{\arraystretch}{1.22}
\begin{tabularx}{\textwidth}{@{}>{\raggedright\arraybackslash}X*{5}{>{\centering\arraybackslash}p{1.85cm}}@{}}
\toprule
\textbf{Benchmark} & \textbf{Concept Grounding} & \textbf{Fact Coverage} & \textbf{Multilingual} & \textbf{Concept Discrimination} & \textbf{Paired Evaluation}\\
\midrule
FiNER \citep{loukas-etal-2022-finer} & \cmark & \cmark  & \xmark & \xmark & \xmark\\
FNXL \citep{sharma-etal-2023-financial} & \cmark  & \cmark  & \xmark & \xmark & \xmark\\
FinTagging \citep{wang2026fintagging} & \cmark  & \cmark  & \xmark & \xmark & \xmark\\
FinQA \citep{chen-etal-2021-finqa} & \xmark & \cmark & \xmark & \xmark & \xmark\\
TAT-QA \citep{zhu-etal-2021-tat} & \xmark & \xmark & \xmark & \xmark & \xmark\\
MultiFin \citep{jorgensen-etal-2023-multifin} & \xmark & \xmark & \cmark  & \xmark & \xmark\\
MultiFinBen \citep{peng-etal-2026-multifinben} & \cmark& \cmark  & \cmark  & \xmark & \xmark\\
XTREME-R \citep{ruder-etal-2021-xtreme} & \xmark & \cmark  & \cmark  & \xmark& \xmark\\
\midrule
\textbf{\bench{} (Ours)} & \cmark  & \cmark  & \cmark  & \cmark  & \cmark \\
\bottomrule
\end{tabularx}
\caption{\textbf{Comparison of benchmarks for financial and multilingual numerical grounding.} Concept grounding evaluates the mapping of numerical mentions to semantic concepts. Fact coverage assesses recovery of annotated evidence, including named entities and supporting facts, rather than only final answers; it need not concern numerical facts. Multilingual indicates evaluation across multiple languages. Concept discrimination tests whether evidence instances express the same or different concepts without candidate concept labels; it does not require an odd-one-out format. Paired evaluation examines the relationship between concept recognition and evidence coverage on matched instances, rather than simply including both tasks. Checks may refer to different tasks within a suite, including FNXL in MultiFinBen; FinQA reports retrieval recall. \ding{51} indicates an explicitly evaluated capability; \ding{55} indicates that the capability is not an explicit evaluation target.}
\label{tab:benchmark-comparison}
\end{table*}

To this end, we introduce \bench{}, built from FY2025 European Single Electronic Format (ESEF) reports published by 90 issuers in English, French and Greek. During report preparation, issuers' financial reporting personnel map disclosed numerical values to concepts in the IFRS Accounting Taxonomy and embed these mappings in the original text and tables through Inline XBRL (iXBRL) tags \citep{esmaElectronic,esma2025manual}. We extract the numerical facts, their contexts and concept annotations, then use shared IFRS concepts to link excerpts across languages and construct paired tasks covering 187 concepts. The excerpts come from different issuers' actual disclosures rather than translated versions of the same passage. To assess annotation quality, two financial experts independently review a stratified sample, with their agreed corrections and exclusions incorporated into the final dataset (\S\ref{sec:review}). Models receive the task text and tables without the original iXBRL tags.

\begin{figure*}[t]
\centering
\includegraphics[width=0.98\linewidth]{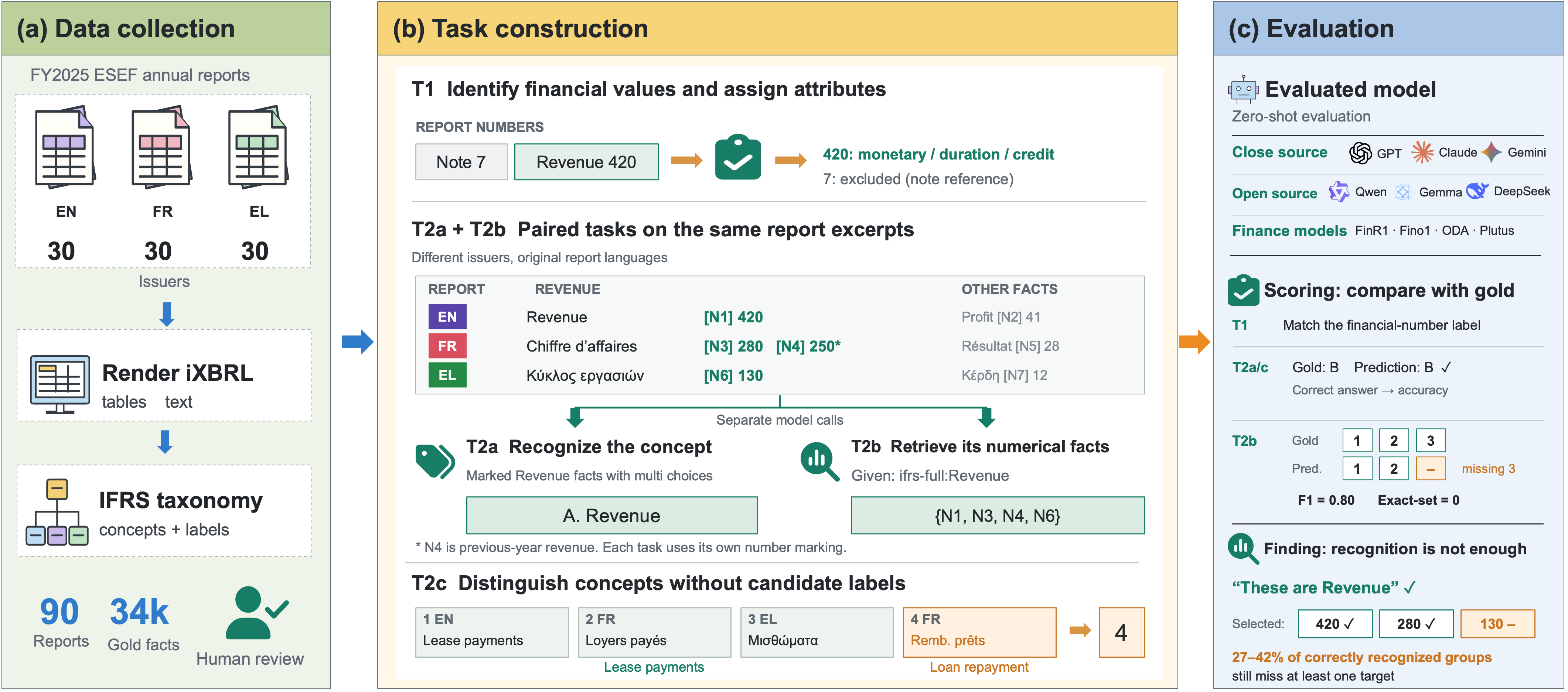}
\caption{\textbf{\bench{} overview.} Reports from 30 issuers per language provide evidence linked through shared accounting concepts. Four independent tasks test Fact Identification (T1), Concept Recognition (T2a), Exhaustive Fact Selection (T2b) and Concept Discrimination without candidate labels (T2c). T2a/T2b use matched contexts to relate recognition to target coverage. Examples are schematic; the finding concerns omissions among correctly recognized groups for the eight reference models. 
}
\label{fig:framework}
\end{figure*}

We design four tasks to distinguish difficulties in numerical interpretation, concept recognition and evidence recovery (Figure~\ref{fig:framework}). \textbf{Fact Identification (T1)} first tests whether models can distinguish financial values from other numbers and identify their accounting attributes. The core paired tasks assess two directions: \textbf{Concept Recognition (T2a)} presents marked numerical facts and asks which accounting concept they share, while \textbf{Exhaustive Fact Selection (T2b)} provides a target concept and asks for all corresponding values in the same contexts. This design evaluates each capability separately and examines whether the same model recognizes a concept correctly yet omits relevant evidence during extraction. Because T2a provides candidate concepts that may help models identify the answer, \textbf{Concept Discrimination (T2c)} additionally asks which excerpt expresses a different concept without providing candidate concept labels. All tasks use independent calls. Diagnostic variants change the evidence languages and taxonomy-label presentation to examine how performance depends on these conditions.

Results show that across 15 models, eight models achieve 97.3--99.4\% recognition accuracy yet miss a target in 26.6--41.9\% of correctly recognized groups. Further, evidence recovery varies by language. Greek has the lowest recall for seven of these eight models, while concept recognition remains high without English evidence. The evaluated finance-specialized models show limited gains over their corresponding base models under our protocol. These findings highlight the need to evaluate cross-language concept recognition alongside the completeness of numerical evidence recovered in each language.

We make three contributions in our work. \textbf{(1)} We construct a benchmark linking real multilingual disclosures through a shared accounting taxonomy and traceable fact annotations. \textbf{(2)} We pair concept recognition with exhaustive fact selection on matched evidence to measure omissions even when the model recognizes the correct concept. T1 and T2c complement this pair by testing financial-fact interpretation and distinctions between related concepts without candidate labels, respectively. \textbf{(3)} Evaluating 15 models reveals weaknesses hidden by recognition scores and shows how performance varies with evidence language, label presentation and financial specialization.

\section{Related Work}
Many existing financial benchmarks evaluate how models link numbers to accounting concepts \citep{loukas-etal-2022-finer,sharma-etal-2023-financial,wang2026fintagging}, and FinTagging also examines errors across extraction stages. FinQA and TAT-QA test reasoning over financial evidence, with FinQA additionally reporting retrieval recall \citep{chen-etal-2021-finqa,zhu-etal-2021-tat}. MultiFin and MultiFinBen evaluate financial tasks across languages \citep{jorgensen-etal-2023-multifin,peng-etal-2026-multifinben}. However, these evaluations do not directly show whether models that recognize a shared accounting concept can also recover all corresponding numerical facts from the same multilingual evidence (Table~\ref{tab:benchmark-comparison}). Appendix~\ref{app:related-work} provides a fuller comparison.

\section{The \bench{} Benchmark}
\label{sec:tasks}
\subsection{Data Collection}
\label{sec:data}
We collected FY2025 ESEF reports from 30 issuers (reporting companies) per language, each appearing in only one language. All cover a full year ending 31 December 2025, with initial fact-to-concept mappings come from issuer-provided iXBRL tags. Appendix Table~\ref{tab:issuers} lists all 90 issuers, their languages and sectors; Appendix~\ref{app:construction} details the English filing index, French CAC~40 and Greek Athens Exchange sources and selection criteria.

\subsection{From iXBRL Reports to Evidence}
We render reports as text and Markdown tables, retaining headers and each fact's offsets, concept and accounting attributes. Appendix~\ref{app:construction} details extraction, normalization and filtering.

ESEF requires individual tags mainly in primary statements; notes use block tags \citep{esma2025manual}. Most individually tagged numbers are amounts, per-share amounts or share counts. Initial annotations come from issuer tags and the taxonomy, with expert corrections or exclusions applied to a sampled subset (\S\ref{sec:review}). Issuer-specific extensions, which add concepts not covered by the standard taxonomy, appear only in Fact Identification (T1). The concept-level tasks (T2a, T2b and T2c) use standard \concept{ifrs-full} concepts. Table~\ref{tab:data} summarizes the data.

\noindent\textbf{Illustrative example.} Consider invented excerpts reporting revenue of EUR 420 million in English, EUR 280 million and 250 million for the current and prior years in French, and EUR 130 million in Greek. Each also contains other figures, such as profit. T1 identifies financial values and their attributes. T2a highlights the revenue figures and asks for their shared concept from candidate labels. T2b provides \emph{Revenue} and asks for all four values, including the French comparative amount. In a separate four-excerpt T2c item, three excerpts mark revenue and one marks profit; the model selects the odd excerpt without candidate concept labels. 
Appendix~\ref{app:example} gives the full example.

\subsection{T1: Fact Identification with Accounting Attributes}
Recovering financial evidence first requires distinguishing financial values from other numbers and interpreting their accounting roles. Therefore, we propose the first task T1, it asks which numbers are financial facts and what accounting roles they have. Each item is a passage or table. Every number-like string receives an identifier; splitting and marking rules are in Appendix~\ref{app:construction}. Models identify financial values and return \emph{type} (monetary, per share, shares), \emph{period type} (instant: a value at a date; duration: a value over a period) and \emph{balance} (the concept's normal debit or credit side for monetary amounts). For a context $x$ with numbered instances $N_x$, the output is
\begin{equation}
f_1(x,N_x)=\{(n,\widehat{\boldsymbol a}_n):n\in\widehat F_x\},
\label{eq:t1}
\end{equation}
where $\widehat F_x\subseteq N_x$ contains selected facts and $\widehat{\boldsymbol a}_n$ is the predicted (type, period, balance) tuple for instance $n$. These are concept attributes, not properties of the displayed sign. Appendix~\ref{app:formalism} gives the full definitions and scoring conditions.

Block-tagged notes may contain untagged financial values. Untagged numbers are scored as negatives only for clearly excluded categories: years, dates, percentages, references, footnote markers, and counts of years, days or employees.

\begin{table}[!t]
\centering\footnotesize\setlength{\tabcolsep}{3.5pt}
\resizebox{\ifdim\width>\columnwidth\columnwidth\else\width\fi}{!}{%
\begin{tabular}{@{}lccc@{}}
\toprule
 & EN & FR & EL\\
\midrule
Issuers (FY2025 annual reports) & 30 & 30 & 30\\
\midrule
\textit{T1} items (tables / text) & 435 (368 / 67) & 450 (368 / 82) & 422 (415 / 7)\\
\quad marked numbers per item & 35.2 & 35.9 & 30.1\\
\quad financial facts (gold) & 11,583 & 12,191 & 10,287\\
\quad\quad monetary, instant, debit & 1,277 & 1,360 & 1,216\\
\quad\quad monetary, instant, credit & 2,408 & 2,502 & 2,311\\
\quad\quad monetary, duration, debit & 2,550 & 2,433 & 2,262\\
\quad\quad monetary, duration, credit & 4,948 & 5,319 & 4,240\\
\quad\quad monetary, no balance (unscored) & 131 & 169 & 130\\
\quad\quad per share / shares & 226 / 43 & 255 / 153 & 120 / 8\\
\quad non-financial numbers (class \emph{none}) & 2,801 & 3,405 & 1,638\\
\quad unscored untagged amounts & 907 & 577 & 760\\
\midrule
\textit{T2a/T2b}: 521 groups, 187 concepts; contexts & 521 & 521 & 521\\
\quad target facts per context & 2.5 & 2.6 & 2.7\\
\quad numbered facts per context (T2b) & 13.9 & 13.8 & 13.0\\
\textit{T2c}: 534 items, 267 concepts; excerpts & 790 & 729 & 617\\
\quad odd excerpt in this language & 221 & 168 & 145\\
\bottomrule
\end{tabular}}
\caption{\textbf{Benchmark statistics.} Financial-fact annotations originate from issuer iXBRL tags, with expert corrections and exclusions incorporated for the audited subset ($\S$\ref{sec:review}). T1 counts financial facts by type, period type and balance; non-financial numbers are scored in class \emph{none}, while untagged amount-like numbers are unscored. }
\label{tab:data}
\end{table}

\subsection{T2a: Concept Recognition}
Comparing financial disclosures across languages requires recognizing when differently worded numerical facts express the same accounting concept. And T2a asks which accounting concept the marked facts share across languages.
For each target concept $c$, we combine English, French and Greek contexts from different issuers. A context is a passage or table window. All occurrences of $c$ are marked (2.5--2.7 per context on average). Of 190 concepts tagged by at least two issuers per language, 187 have three distractors that meet our criteria. Each yields up to three groups (521 total). Let $X$ denote the contexts, $H$ is the marked target instances, and $C$ is the four candidate concepts. The model predicts
\begin{equation}
\widehat c=f_a(X,H,C)\in C.
\label{eq:t2a}
\end{equation}
Each option shows a \concept{ifrs-full:Concept} identifier and English label. Distractors share the target's datatype and presentation section; selection and exclusion rules appear in Appendix~\ref{app:construction}.

\subsection{T2b: Exhaustive Fact Selection}
T2b reverses the question: given a concept, can the model find every corresponding fact?
Using the same contexts, we number every \concept{ifrs-full} fact across the group (40.6 on average). Given the target concept and its English label, models select all its occurrences, including comparative-period values, and reject other facts (7.8 targets per group). Writing $U_X$ for all numbered facts and $c$ for the gold target concept, the prediction is
\begin{equation}
\widehat S=f_b(X,U_X,c)\subseteq U_X.
\label{eq:t2b}
\end{equation}
The gold set is $S^\star=\{n\in U_X:\tau(n)=c\}$, where $\tau$ is the gold concept assignment. T2a and T2b use separate, stateless calls; T2b receives $c$, not $\widehat c$.

\subsection{T2c: Concept Discrimination}
T2c asks which excerpt expresses a different concept without candidate concept labels.
T2c presents four excerpts from different issuers, covering all three languages with one repeated. Three excerpts mark facts of concept $c$; the fourth marks a confusable concept $d$ with the same datatype and period type. Appendix~\ref{app:construction} gives the confusability and exclusion rules. Let $E=((x_i,H_i))_{i=1}^{4}$ collect the excerpts and their marked instances. The model selects the odd excerpt (chance 25\%):
\begin{equation}
\widehat j=f_c(E)\in\{1,2,3,4\}.
\label{eq:t2c}
\end{equation}
Concept labels are not provided. Concepts present in all three languages yield up to two items with different odd concepts: 534 items over 267 concepts and 303 odd concepts. We rotate the odd excerpt's language and shuffle its position. In 275 items, another excerpt shares its language, preventing language difference from identifying the answer.

\subsection{Taxonomy Conditions and Diagnostic Variants}
\label{sec:conditions}
\textbf{C1} (primary) shows options as concept identifier plus English label. \textbf{C2} shows only the official ESMA French or Greek label (no identifier) on French-only or Greek-only items, removing English from the options while retaining the shared English instructions. \textbf{C3} shows English, French and Greek labels on joint items. All T2a variants use the same items, options and option order. We also run T2a and T2b without the English excerpt, and T2a with only one language. Two harder option sets (Appendix~\ref{app:results}) keep the T2a items but allow calculation siblings and rank distractors by word overlap with the target's typical row labels, with four or twenty options.

\subsection{Expert Validation}
\label{sec:review}
Two financial experts independently reviewed a stratified 10\% sample per task: 131 T1 items, 52 groups shared by T2a/T2b, and 53 T2c items. Sampling covers language, format, data type and item size, with at most one group per concept. Reviewers checked annotations against IFRS definitions, then resolved disagreements through discussion. Appendix~\ref{app:review} details qualifications, sampling, guidelines and uncertainty estimates.

Corrections and exclusions apply to the reviewed subset; other items retain automatically derived annotations. All reported benchmark results use this combined final annotation set.

Agreement before discussion is $\kappa=\target{0.87}$ (T1), \target{0.74} (T2a), \target{0.84} (T2b) and \target{0.88} (T2c). The audit corrects \target{1.3}\% of T1 numbers and \target{2.1}\% of T2b facts, finds \target{5.1}\% of T2a distractors ambiguous, and confirms the odd excerpt for \target{96.2}\% of audited T2c items (Table~\ref{tab:review}). Replacing original with corrected annotations changes reference-model scores by at most \target{1.4} percentage points on audited subsets. This measures sensitivity to identified errors, not correctness of unreviewed annotations.

\section{Experimental Setup}
\label{sec:setup}
We evaluate 15 models in four groups (Appendix Table~\ref{tab:models}). The API models are GPT-5.6 Luna \citep{openai2026catalog}, Claude Haiku 4.5 \citep{anthropic2025haiku}, Gemini 3.8 Flash \citep{google2026gemini} and DeepSeek-V4.1-Flash \citep{deepseek2026v41}. Open-weight references are Qwen3.8-27B \citep{qwen2026open}, Qwen3.8-Flash-Next \citep{qwen2026next}, Gemma 4 31B IT \citep{gemma2026report} and Qwen3.5-122B-A10B \citep{qwen35card2026}. Finance specialists are Fin-R1 \citep{liu2025finr1}, Fin-o1-8B \citep{qian2025fino1}, ODA-Fin-RL-8B \citep{cao2026odafin} and Greek Plutus-8B-Instruct \citep{peng-etal-2025-plutus}. Their bases are Qwen2.5-7B-Instruct \citep{qwen2024report}, Qwen3-8B \citep{yang2025qwen3} and Llama-Krikri-8B-Instruct \citep{roussis-etal-2025-krikri}.

Each call contains English instructions and original-language evidence, without a system prompt, tools, retrieval or history. Prompts request JSON (T1/T2b), a letter (T2a) or an excerpt number (T2c); see Appendix~\ref{app:prompts}. They average 1.2k input tokens (maximum 3.7k). The output budget is 16,384 tokens including reasoning. We use each model's default decoding settings and disable API-model reasoning. Separate reasoning channels or thinking tags are removed before scoring. The task and scoring protocols are shared, but reasoning and generation settings do not equalize compute (Appendix~\ref{app:model-manifest}).

All results use final annotations incorporating expert corrections to audited items. Only the audit sensitivity analysis compares original and corrected annotations.

\subsection{Evaluation Metrics}
\label{sec:metrics}

For \textbf{T1}, scored numbers belong to seven classes: monetary amounts by period and balance (four), per-share amounts, shares, and \emph{none}. Omitted numbers receive \emph{none}; unscored numbers are ignored. We pool per-class scores across items and report seven-class macro-F1. Joint fact F1 requires every attribute to be correct; period and balance accuracy are conditional on correct fact identification. For \textbf{T2a and T2c}, we report accuracy of the chosen option or excerpt.
For \textbf{T2b}, we compute precision, recall and F1 against each group's target set. Recall measures the proportion of target facts recovered; precision measures the proportion of selected facts that belong to the target set. The primary score averages group F1 within each concept, then across concepts (concept-macro F1). Complete target coverage requires all target facts to be selected but permits additional incorrect selections. Exact-set accuracy requires the predicted set to match the target set exactly, with neither omissions nor extra selections. Per-language precision, recall and F1 pool mentions within each language.

\noindent\textbf{Strict protocol.} An output is invalid if the request failed, the budget ran out before an answer, the model refused, or the answer does not follow the task format. Invalid outputs are scored as wrong (T1: no predictions; T2a, T2c: incorrect; T2b: F1 and exact match 0). We report invalid rates alongside every score.

\noindent\textbf{Uncertainty.} T1 intervals resample issuers within language (2,000 bootstrap samples). T2 groups reuse the 30 issuers per language. The graph linking groups to issuers forms a single connected component, so we do not report bootstrap intervals for T2. We use exact McNemar tests for paired comparisons of conditions. These item-level $p$-values do not account for dependence between items that share issuers. We therefore treat them as exploratory.

\noindent\textbf{Baselines.} We use two simple baselines. The random baseline samples a valid answer uniformly at random. The majority/select-all baseline assigns every number to the most frequent T1 class: monetary amounts with duration period type and credit balance. For T2a and T2c, it always selects the first option or excerpt. For T2b, it selects every numbered fact, ensuring complete target coverage while also including non-target facts.

\section{Results}
\label{sec:results}
\begin{table*}[!t]
\centering\footnotesize\setlength{\tabcolsep}{3.5pt}
\resizebox{\ifdim\width>\textwidth\textwidth\else\width\fi}{!}{%
\begin{tabular}{@{}lcccccccccc@{}}
\toprule
 & \multicolumn{5}{c}{T1 (7-class macro-F1)} & T2a & \multicolumn{2}{c}{T2b} & T2c & Invalid (\%)\\
\cmidrule(lr){2-6}\cmidrule(lr){8-9}
Model & All [95\% CI] & EN & FR & EL & Text & Acc. & Macro-F1 & Exact set & Acc. & T1/T2a/T2b/T2c\\
\midrule
\multicolumn{11}{@{}l}{\textit{API models}}\\
GPT-5.6 Luna & 88.6\,{\scriptsize[87.3, 89.5]} & 87.2 & 88.5 & 84.7 & 73.6 & 97.3 & 88.3 & 65.8 & 49.4 & 0/0/0/0\\
Claude Haiku 4.5 & 85.9\,{\scriptsize[84.4, 87.0]} & 84.6 & 86.4 & 83.7 & 71.9 & 97.5 & 86.2 & 60.1 & 59.9 & 0/0/0/0\\
Gemini 3.8 Flash & 95.4\,{\scriptsize[94.5, 96.2]} & 95.9 & 95.1 & 95.1 & 86.9 & 99.4 & 89.2 & 65.6 & 95.1 & 1/0/0/0\\
DeepSeek-V4.1-Flash & 88.1\,{\scriptsize[87.1, 88.9]} & 87.7 & 86.9 & 88.1 & 72.4 & 98.3 & 86.2 & 59.7 & 35.6 & 0/0/0/8\\
\midrule
\multicolumn{11}{@{}l}{\textit{Open-weight models}}\\
Qwen3.8-27B & 68.1\,{\scriptsize[65.1, 70.9]} & 72.3 & 63.4 & 75.5 & 86.5 & 98.5 & 62.5 & 57.0 & 72.3 & 31/1/34/27\\
Qwen3.8-Flash-Next & 66.6\,{\scriptsize[63.7, 69.4]} & 69.7 & 62.1 & 74.1 & 84.1 & 98.5 & 73.7 & 67.4 & 75.5 & 33/2/23/23\\
Gemma 4 31B IT & 90.5\,{\scriptsize[89.2, 91.7]} & 90.6 & 89.2 & 93.1 & 75.4 & 98.8 & 87.4 & 65.1 & 89.0 & 0/0/0/0\\
Qwen3.5-122B-A10B & 87.6\,{\scriptsize[86.1, 88.8]} & 87.8 & 86.2 & 85.9 & 83.5 & 99.0 & 90.4 & 71.8 & 91.6 & 6/0/0/1\\
\midrule
\multicolumn{11}{@{}l}{\textit{Finance models}}\\
Fin-R1 & 31.1\,{\scriptsize[29.0, 33.1]} & 31.1 & 35.5 & 24.2 & 39.0 & 82.7 & 66.8 & 23.4 & 27.5 & 0/0/2/0\\
Fin-o1-8B & 51.0\,{\scriptsize[48.4, 53.5]} & 53.4 & 52.3 & 43.5 & 50.6 & 87.3 & 74.3 & 39.0 & 47.4 & 0/3/0/0\\
ODA-Fin-RL-8B & 69.8\,{\scriptsize[67.1, 72.2]} & 70.2 & 73.0 & 61.3 & 68.3 & 90.6 & 81.1 & 49.7 & 45.7 & 1/7/1/43\\
Plutus-8B-Instruct & 12.6\,{\scriptsize[10.5, 14.8]} & 11.8 & 16.0 & 10.7 & 22.6 & 77.5 & 10.7 & 0.2 & 27.2 & 80/0/71/0\\
\midrule
\multicolumn{11}{@{}l}{\textit{Base models of the finance models}}\\
Qwen2.5-7B-Instruct & 39.5\,{\scriptsize[37.2, 41.7]} & 43.2 & 41.3 & 31.2 & 50.0 & 88.3 & 57.0 & 7.5 & 30.5 & 0/0/0/0\\
Qwen3-8B & 70.7\,{\scriptsize[67.9, 73.3]} & 72.3 & 73.1 & 61.9 & 62.0 & 94.2 & 80.4 & 45.5 & 68.7 & 0/0/0/0\\
Llama-Krikri-8B-Instruct & 37.3\,{\scriptsize[35.8, 38.8]} & 39.4 & 35.6 & 36.1 & 34.7 & 84.1 & 44.9 & 1.0 & 27.3 & 7/0/0/0\\
\midrule
\multicolumn{11}{@{}l}{\textit{Baselines}}\\
Random & 8.8 & 9.4 & 8.7 & 8.0 & 10.3 & 23.8 & 27.5 & 0.0 & 24.7 & 0/0/0/0\\
Majority / select-all & 7.4 & 7.4 & 7.3 & 7.6 & 2.5 & 24.2 & 32.9 & 0.0 & 24.9 & 0/0/0/0\\
\bottomrule
\end{tabular}}
\caption{\textbf{Main results} (\%, strict protocol: invalid outputs count as wrong). T1 intervals resample issuers within language (2,000 bootstrap samples); T2 items share issuers across groups, so no T2 intervals are reported (Section~\ref{sec:setup}). Text: T1 text passages only.}
\label{tab:main}
\end{table*}

Table~\ref{tab:main} reports all 15 models under strict scoring. The eight API/open-weight \emph{reference models} achieve 97.3--99.4\% T2a accuracy, yet the best T2b scores are 90.4 concept-macro F1 and 71.8\% exact-set accuracy. T2c accuracy ranges from 35.6--95.1\%, revealing variation not apparent from recognition accuracy alone (Appendix~\ref{app:score-details}).

\subsection{Does Recognition Imply Coverage?}
\label{sec:rq1}
\begin{figure}[!t]
\centering
\includegraphics[width=0.8\linewidth]{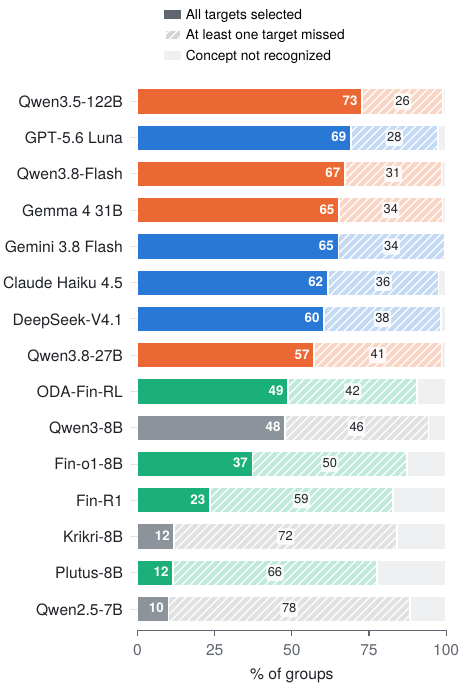}
\caption{\textbf{Recognition and target coverage} over 521 groups. The first two categories require correct T2a recognition; coverage concerns all three languages in T2b. Selecting all targets permits extra selections and is not exact-set correctness. Colours: API (blue), open-weight (orange), finance (green), base (grey). 
}
\label{fig:grounding}
\end{figure}
We examine whether correctly answered T2a groups also yield complete target coverage in T2b (Figure~\ref{fig:grounding}; Table~\ref{tab:paired}). Despite 97--99\% recognition accuracy, reference models miss a target in at least one language in 26.6\% (Qwen3.5-122B) to 41.9\% (Qwen3.8-27B) of these groups. Precision remains high at 88.5--100\% per language, while recall falls to 51.6--83.5\%. Greek has the lowest recall for seven of the eight reference models. Gemini's omission rate rises from 27.7\% for recognized groups with 3--6 targets to 46.3\% for those with 7 or more. Omissions are more common in groups with more targets, although this comparison does not isolate the effect of target count. Finance models and their bases show larger gaps (Appendix~\ref{app:score-details}). Selecting every fact guarantees target coverage, so coverage must be read alongside precision and exact-set accuracy, which penalize extra selections. The paired results nevertheless show frequent omissions for correctly recognized concepts, revealing incomplete evidence recovery that recognition accuracy alone cannot expose.

\subsection{How Does Evidence Language Affect Grounding?}
\label{sec:rq2}
\begin{figure}[!t]
\centering
\includegraphics[width=0.8\linewidth]{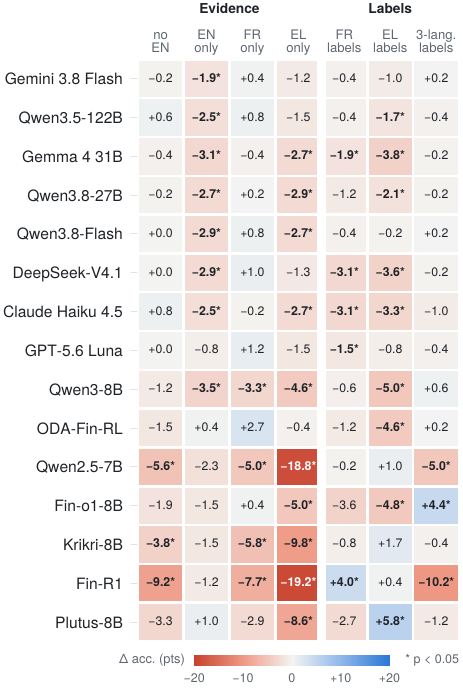}
\caption{\textbf{T2a changes across evidence and label conditions} (points). Evidence conditions compare with joint inputs; C2 compares localized labels without identifiers with C1's English labels and identifiers; C3 adds multilingual labels. C2 does not isolate label language. * McNemar $p<0.05$. Scores: Table~\ref{tab:language}.}
\label{fig:language}
\end{figure}
\begin{figure}[!t]
\centering
\includegraphics[width=0.8\linewidth]{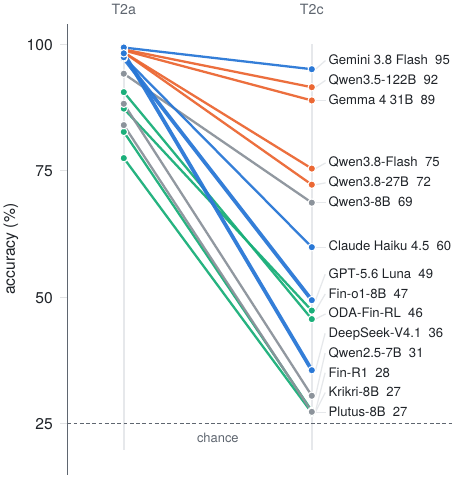}
\caption{\textbf{From recognition to discrimination.} T2a accuracy (shared concept, four labelled options) and T2c accuracy (odd one out, no candidate labels) of each model; dashed line: chance. The tasks also differ in evidence and structure, so the score gap does not isolate the effect of candidate labels.}
\label{fig:t2a-t2c}
\end{figure}
For reference models, the difference in T2a accuracy between joint inputs and inputs without English is $-$0.8 to $+$0.4 points (Figure~\ref{fig:language}). Recognition remains high without the English excerpt on these items. Finance models and their bases are more sensitive. Fin-R1 loses 9.2 points without English, Qwen2.5-7B 5.6, Llama-Krikri-8B 3.8 and Plutus 3.3. Keeping only English is more costly than removing it: seven of the eight reference models score 1.9--3.1 points below the joint condition (McNemar $p<0.05$; GPT-5.6 Luna $-$0.8, not significant), whereas French-only scores stay within 1.2 points. These comparisons do not isolate language competence because the issuers and contexts differ. T2b effects vary by model and language; fewer budget failures partly explain Qwen3.8 gains on shorter prompts (Appendix~\ref{app:score-details}).

T2c extends this analysis to language configurations in concept discrimination and reveals variation not apparent from T2a accuracy (Figure~\ref{fig:t2a-t2c}). Gemini scores 95.1\%, Qwen3.5-122B 91.6\% and Gemma 89.0\%, while DeepSeek scores 98.3\% on T2a but 35.6\% on T2c. Every model clearly above chance is less accurate when another excerpt shares the odd excerpt's language. Accuracy therefore varies with the language configuration, although this comparison does not establish which cues the models use. 


\subsection{How Does Taxonomy Label Presentation Affect Recognition?}
\label{sec:rq3}
On Greek-only items, C2 significantly lowers accuracy for eight models (Figure~\ref{fig:language}), by 1.7--5.0 points. Only Greek-adapted Plutus improves significantly ($+$5.8, $p=0.007$). French-label effects are smaller and vary by model. C3 changes reference-model scores by at most 1.0 point. None of these changes is significant. Full differences and McNemar $p$-values appear in Appendix~\ref{app:label-details}. Localized taxonomy labels do not consistently improve recognition under these conditions. 

\subsection{What Makes T1 Hard?}
\label{sec:t1}
Conditional on correct fact identification, reference models achieve at least 99.9\% type accuracy and 98.4\% period accuracy (Table~\ref{tab:t1-detail}, Figure~\ref{fig:t1-classes}). Balance accuracy is lower at 80.6--95.7\%, and debit-balance flows are the lowest-scoring class. Most models perform worse on passages than tables, although the two formats differ in content. Qwen3.8 shows the reverse pattern: budget exhaustion affects 31--33\% of T1 items, mostly long tables. Grounding thus requires interpreting accounting roles as well as locating values (Appendix~\ref{app:score-details}).

\subsection{Does Financial Specialization Help?}
\label{sec:finance}
In our evaluation, all four finance-specialized models perform worse than their base models on T1, T2a, and T2c (Figure~\ref{fig:finance}). Fin-R1 improves on T2b by 9.9 macro-F1 points, with larger gains in French and Greek of 12.5 and 14.8 F1 points, respectively. Plutus, despite its Greek specialization, scores 25.4 points below its base model on Greek T1. Its high failure rate partly explains this result: 80\% of its T1 outputs are invalid, most of them empty. Overall, we find limited benefits from these finance-specialized models under our evaluation settings for capturing concepts and corresponding numbers. 

\subsection{Robustness to Harder Options and Output Failures}
The lexical-overlap baseline scores 70.2\% on T2a. Harder distractors reduce it to 49.1\% with four options and 36.3\% with twenty, while reference models retain 94.4 to 98.7\% and 89.4 to 97.5\% (Table~\ref{tab:t2a-options}). Strong T2a performance is therefore not explained by this simple word-matching baseline alone, although other wording cues may remain.

At last, we find out that generation failures also affect rankings: Qwen3.8 exhausts its budget on 23--34\% of T1/T2b/T2c items, ODA-Fin-RL-8B on 43\% of T2c, and Plutus frequently returns empty answers. The strict scoring retains these generation failures.

\section{Conclusion}
We introduced \bench{} to examine whether models can recognize shared accounting concepts and recover all corresponding numerical facts across English, French, and Greek reports. Our results show that models often identify the correct concept while missing relevant values. Evidence recovery also varies by language, with Greek recall lowest for seven of the eight reference models. Models continue to struggle with accounting attributes and distinctions between related concepts, while the evaluated finance-specialized models offer limited improvements. These findings show why multilingual financial extraction should consider not only whether models recognize concepts, but also whether they find all the evidence.

\section*{Limitations}
This paper has limitations in annotation quality, data coverage, and evaluation. Annotations come from issuer-provided iXBRL tags, which may contain errors or inconsistent concept choices. Financial experts reviewed stratified samples covering 10\% of each task, and all reported results incorporate their corrections and exclusions. Errors and ambiguities may remain in the unreviewed portion. Also, T1 leaves untagged amount-like numbers unscored. Most T2 contexts are table windows, and concepts shared across all three languages tend to be common primary-statement concepts. The benchmark covers three languages within one reporting regime. Since language varies together with issuer, sector, and document style, differences between single-language conditions cannot be attributed to language alone. 

\section*{Ethical Considerations}
The benchmark uses public regulatory filings and the IFRS Accounting Taxonomy. We document the access conditions for each source. Where full documents cannot be redistributed, the release provides annotations, source information and instructions for reconstructing the inputs. Reports contain company information only; no personal data are part of the benchmark inputs. The two financial experts who reviewed the annotations are co-authors of this paper. Benchmark scores do not establish suitability for financial reporting compliance and do not replace professional review.

\section*{Use of AI Assistants}

The authors used AI assistants, including ChatGPT, to support language polishing, brainstorming, and LaTeX editing. All technical claims, experimental results, analysis, and final writing decisions were reviewed and verified by the authors.

\bibliographystyle{plainnat}
\bibliography{references}

\appendix
\clubpenalty=10000
\widowpenalty=10000

\section{Formal Task Definitions}
\label{app:formalism}
Equations~\ref{eq:t1}--\ref{eq:t2c} describe independent model calls. Gold assignments use final annotations and are withheld unless specified.

\noindent\textbf{Notation.} A context $x$ contains numbered instances $N_x$. Equal values at different locations remain distinct instances. Hats denote predictions; $\tau(n)$ is the gold concept. In the joint setting, $\mathcal L=\{\mathrm{EN},\mathrm{FR},\mathrm{EL}\}$ and $X=(x_\ell)_{\ell\in\mathcal L}$.

\noindent\textbf{T1: facts and attributes.} The model selects $\widehat F_x\subseteq N_x$ and assigns each instance the attribute tuple $\widehat{\boldsymbol a}_n=(\widehat t_n,\widehat p_n,\widehat b_n)$. The attribute domains are:
\begin{align*}
t_n&\in\{\text{monetary},\text{per-share},\text{shares}\},\\
p_n&\in\{\text{instant},\text{duration}\},\\
b_n&\in\{\text{debit},\text{credit},\text{none}\}.
\end{align*}
Scored monetary facts have debit or credit balance; per-share amounts and shares have balance \emph{none}. Attributes follow the concept: inventory is a debit-balance stock, whereas a payment for property is a credit-balance flow.

Scoring covers only $D_x\subseteq N_x$. Unselected scored instances receive the non-financial class \emph{none}, distinct from balance \emph{none}. Joint correctness requires the identifier and all three attributes to match.

\noindent\textbf{T2a: shared-concept recognition.} Let $H_\ell$ contain the marked targets in $x_\ell$ and $H=(H_\ell)_{\ell\in\mathcal L}$. For gold concept $c$,
\begin{equation}
\tau(n)=c\quad\text{for every }n\in H_\ell,\ \ell\in\mathcal L.
\end{equation}
The candidate set $C$ contains $c$ and the distractors. The model selects through an option letter, with correctness $\mathbf{1}[\widehat c=c]$; $|C|=4$ in the primary setting.

\noindent\textbf{T2b: exhaustive fact selection.} T2b numbers all standard IFRS facts $U_X$ in the T2a contexts and supplies the gold concept $c$ with its English label. The target set is
\begin{equation}
S^\star(X,c)=\{n\in U_X:\tau(n)=c\}.
\end{equation}
For a valid prediction $\widehat S$, precision and recall are
\begin{equation}
P=\frac{|\widehat S\cap S^\star|}{|\widehat S|},\qquad
R=\frac{|\widehat S\cap S^\star|}{|S^\star|}.
\end{equation}
An empty prediction has zero precision, recall and F1; target sets are nonempty. Group F1 is averaged within concepts, then across concepts. The two set-level conditions are
\begin{align}
\mathrm{Coverage}&=\mathbf{1}[S^\star\subseteq\widehat S],\\
\mathrm{Exact}&=\mathbf{1}[S^\star=\widehat S].
\end{align}
Coverage permits extra selections; exact-set correctness does not. The missing-target rate among correctly recognized groups $g$ is
\begin{equation}
\frac{\sum_g\mathbf{1}[\widehat c_g=c_g]\,\mathbf{1}[S_g^\star\nsubseteq\widehat S_g]}
{\sum_g\mathbf{1}[\widehat c_g=c_g]},
\end{equation}
when the denominator is nonzero. This is the conditional omission rate, not the rate of all non-exact predictions. Invalid outputs follow Section~\ref{sec:metrics}.

\noindent\textbf{T2c: concept discrimination.} The input $E=((x_i,H_i))_{i=1}^{4}$ contains four excerpts. Each $x_i$ marks instances $H_i$ of gold concept $z_i$. A unique odd index $j^\star$ satisfies
\begin{equation}
z_i=\begin{cases}
d,&i=j^\star,\\
c,&i\ne j^\star,
\end{cases}\qquad c\ne d.
\end{equation}
The model sees neither candidate concept labels nor $z_i$ and returns an excerpt index, with correctness $\mathbf{1}[\widehat j=j^\star]$.

\section{Prompts}
\label{app:prompts}
All prompts are in English, followed by the evidence in its original language, with tables in Markdown and the number conventions of each language. Abridged instructions:
\begin{quote}\small
\textbf{T1:} Every number is marked with an identifier ([N1], [N2], \dots). For each marked number that reports a financial value, give its Type (monetaryItemType, perShareItemType, sharesItemType), Period (instant, duration) and Balance (debit, credit; none for per-share amounts and shares), with definitions and examples. Years, dates, references, percentages and other counts are not financial entities. Output only \texttt{\{"result": [\{"ID", "Type", "Period", "Balance"\}]\}}, without explanations.\par
\textbf{T2a:} In the companies' iXBRL filings, all marked numbers across the excerpts are tagged with the same IFRS concept. Which concept is shared by [N1], \dots? Options A--D. Answer with the letter of the correct option only. Output only the letter (e.g.\ B), without explanation.\par
\textbf{T2b:} Target concept: \concept{ifrs-full:Concept}, English label. Select all marked numbers, in all excerpts, that are tagged with the target concept; several may qualify (e.g.\ current and prior year); do not select numbers of other concepts, even closely related ones. Output \texttt{\{"result": ["N1", \dots]\}} without explanations.\par
\textbf{T2c:} The marked numbers of each excerpt are tagged with one IFRS concept; in three excerpts this is the same concept, in one a different concept. In which excerpt are the marked numbers tagged with a different concept? Output only the number (e.g.\ 2), without explanation.
\end{quote}

\section{Models and Runs}
\label{app:model-manifest}
\begin{table*}[t]
\centering\small\setlength{\tabcolsep}{5pt}
\begin{tabularx}{\textwidth}{@{}l l X l@{}}
\toprule
Model & Group & Access & Reasoning in our runs\\
\midrule
GPT-5.6 Luna & API & OpenAI \path{gpt-5.6-luna} (via Azure OpenAI) & off\\
Claude Haiku 4.5 & API & Anthropic \path{claude-haiku-4-5} (via Microsoft Foundry) & off\\
Gemini 3.8 Flash & API & Google \path{gemini-3.8-flash} & off\\
DeepSeek-V4.1-Flash & API & DeepSeek \path{deepseek-flash} & off\\
\midrule
Qwen3.8-27B & Open weights & \path{Qwen/Qwen3.8-27B} & model default (thinks)\\
Qwen3.8-Flash-Next & Open weights & \path{Qwen/Qwen3.8-Flash-Next} & model default (thinks)\\
Gemma 4 31B IT & Open weights & \path{google/gemma-4-31B-it} & model default (no thinking)\\
Qwen3.5-122B-A10B & Open weights & \path{Qwen/Qwen3.5-122B-A10B} (MoE, 10B active) & model default (thinks)\\
\midrule
Fin-R1 & Finance & \path{SUFE-AIFLM-Lab/Fin-R1} (base: Qwen2.5-7B-Instruct) & trained reasoning\\
Fin-o1-8B & Finance & \path{TheFinAI/Fin-o1-8B} (base: Qwen3-8B) & trained reasoning\\
ODA-Fin-RL-8B & Finance & \path{OpenDataArena/ODA-Fin-RL-8B} (base: Qwen3-8B) & trained reasoning\\
Plutus-8B-Instruct & Finance (Greek) & \path{TheFinAI/plutus-8B-instruct} (base: Llama-Krikri-8B) & none\\
\midrule
Qwen2.5-7B-Instruct & Base model & \path{Qwen/Qwen2.5-7B-Instruct} & none\\
Qwen3-8B & Base model & \path{Qwen/Qwen3-8B} & model default (thinks)\\
Llama-Krikri-8B-Instruct & Base model & \path{ilsp/Llama-Krikri-8B-Instruct} & none\\
\bottomrule
\end{tabularx}
\caption{\textbf{Evaluated models}: four API models (one per provider family), four open-weight models, four finance models and the three base models they were adapted from. Open weights are served locally with SGLang in their default chat template and generation configuration; API models run with reasoning switched off. Every model receives the same prompts and an output budget of 16,384 tokens including reasoning.}
\label{tab:models}
\end{table*}

Table~\ref{tab:models} lists the evaluated models. We query Gemini through Google's OpenAI-compatible endpoint and DeepSeek through its own endpoint. GPT-5.6 Luna and Claude Haiku 4.5 are accessed through Azure OpenAI and Microsoft Foundry deployments, using the OpenAI and Anthropic SDKs. For every response, we record the model name returned by the provider and the request identifier. All four API models run with reasoning switched off: \texttt{reasoning\_effort=none} for GPT-5.6 Luna and Gemini 3.8 Flash, no \texttt{thinking} parameter for Claude Haiku 4.5 (extended thinking off), and thinking disabled for DeepSeek-V4.1-Flash. Collaborators ran GPT-5.6 Luna and Claude Haiku 4.5 with the same code and data release. Their outputs were scored using the same scorer as all other models. For open-weight models, we disable prefix caching on the Gemma 4 31B IT SGLang server. This avoids a memory-pool failure in this SGLang version when serving Gemma 4. All other servers use default settings apart from the reasoning parser.

\noindent\textbf{Environment.} API models were queried in October 2026 through the providers' SDKs (OpenAI Python SDK 2.6.1, Anthropic SDK 0.121.0) with eight concurrent requests. Open-weight models were served with SGLang 0.5.18 (0.5.21 for Qwen3.8-Flash-Next, whose architecture requires it) on NVIDIA B200 GPUs: one GPU for each model up to 31B parameters, tensor parallelism over two GPUs for Qwen3.5-122B-A10B and over four for Qwen3.8-Flash-Next. Each server uses the model's default chat template and a 32,768-token context. We use a reasoning parser for models that emit thinking tags. Each server receives 64 concurrent requests. Every run records the request configuration, the returned model name, request identifiers, data hashes and the code version.

\section{Additional Results}
\label{app:results}
\subsection{Additional Score Comparisons}
\label{app:score-details}
\begin{table}[!t]
\centering\footnotesize\setlength{\tabcolsep}{3.5pt}
\resizebox{\ifdim\width>\columnwidth\columnwidth\else\width\fi}{!}{%
\begin{tabular}{@{}lccccc@{}}
\toprule
 & T2a & Miss $\geq$1 & \multicolumn{3}{c}{T2b recall}\\
\cmidrule(lr){4-6}
Model & correct & language & EN & FR & EL\\
\midrule
\multicolumn{6}{@{}l}{\textit{API models}}\\
GPT-5.6 Luna & 97.3 & 29.0 & 82.1 & 83.5 & 80.3\\
Claude Haiku 4.5 & 97.5 & 36.8 & 77.7 & 78.7 & 77.1\\
Gemini 3.8 Flash & 99.4 & 34.6 & 81.7 & 82.2 & 75.1\\
DeepSeek-V4.1-Flash & 98.3 & 38.5 & 74.2 & 76.7 & 73.4\\
\midrule
\multicolumn{6}{@{}l}{\textit{Open-weight models}}\\
Qwen3.8-27B & 98.5 & 41.9 & 53.2 & 54.0 & 51.6\\
Qwen3.8-Flash-Next & 98.5 & 31.8 & 60.2 & 62.5 & 61.4\\
Gemma 4 31B IT & 98.8 & 34.0 & 74.9 & 76.9 & 68.9\\
Qwen3.5-122B-A10B & 99.0 & 26.6 & 79.6 & 80.9 & 77.7\\
\midrule
\multicolumn{6}{@{}l}{\textit{Finance models}}\\
Fin-R1 & 82.7 & 71.7 & 70.7 & 61.9 & 47.7\\
Fin-o1-8B & 87.3 & 57.4 & 69.5 & 62.7 & 55.9\\
ODA-Fin-RL-8B & 90.6 & 46.2 & 74.0 & 68.7 & 63.2\\
Plutus-8B-Instruct & 77.5 & 85.1 & 24.8 & 15.5 & 16.7\\
\midrule
\multicolumn{6}{@{}l}{\textit{Base models of the finance models}}\\
Qwen2.5-7B-Instruct & 88.3 & 88.7 & 62.0 & 45.7 & 30.7\\
Qwen3-8B & 94.2 & 49.3 & 75.3 & 70.8 & 64.9\\
Llama-Krikri-8B-Instruct & 84.1 & 85.8 & 87.9 & 41.6 & 30.2\\
\midrule
\multicolumn{6}{@{}l}{\textit{Baselines}}\\
Random & 23.8 & 100.0 & 47.6 & 52.6 & 49.7\\
Majority / select-all & 24.2 & 0.0 & 100.0 & 100.0 & 100.0\\
\bottomrule
\end{tabular}}
\caption{\textbf{Recognition versus grounding} (\%). Among groups whose T2a concept is answered correctly, the share whose T2b selection misses at least one target number in some language, and T2b recall per language on those groups.}
\label{tab:paired}
\end{table}

\noindent\textbf{Overall scores.}
Gemini 3.8 Flash leads T1 with 95.4 macro-F1, followed by Gemma 4 31B (90.5), GPT-5.6 Luna (88.6), DeepSeek-V4.1-Flash (88.1), Qwen3.5-122B (87.6) and Claude Haiku 4.5 (85.9); the 8B models range from 12.6 to 70.7. Finance models and their bases have T2a scores of 77.5--94.2\%. Finance models, Qwen2.5-7B and Llama-Krikri-8B reach 27.2--47.4\% on T2c.

\noindent\textbf{Recognition and recall.}
On T2a-correct groups, Gemini's T2b recall is 81.7 / 82.2 / 75.1\% in English / French / Greek. Qwen2.5-7B recognizes 88.3\% of groups but misses a target in 88.7\% of those groups, with Greek recall of 30.7\%.

\begin{table*}[!t]
\centering\footnotesize\setlength{\tabcolsep}{3.5pt}
\resizebox{\ifdim\width>\textwidth\textwidth\else\width\fi}{!}{%
\begin{tabular}{@{}lcccccccccccc@{}}
\toprule
 & \multicolumn{6}{c}{T2a evidence languages (C1)} & \multicolumn{3}{c}{T2a option labels} & \multicolumn{3}{c}{T2c}\\
\cmidrule(lr){2-7}\cmidrule(lr){8-10}\cmidrule(lr){11-13}
Model & Joint & No EN & EN & FR & EL & $\Delta_\text{EN}$ & FR (C2) & EL (C2) & Joint (C3) & Acc. & Same lang. & Unique\\
\midrule
\multicolumn{13}{@{}l}{\textit{API models}}\\
GPT-5.6 Luna & 97.3 & 97.3 & 96.5 & 98.5 & 95.8 & +0.0 & 96.9 & 95.0 & 96.9 & 49.4 & 46.9 & 52.1\\
Claude Haiku 4.5 & 97.5 & 98.3 & 95.0 & 97.3 & 94.8 & $-$0.8 & 94.2 & 91.6 & 96.5 & 59.9 & 57.8 & 62.2\\
Gemini 3.8 Flash & 99.4 & 99.2 & 97.5 & 99.8 & 98.3 & +0.2 & 99.4 & 97.3 & 99.6 & 95.1 & 93.1 & 97.3\\
DeepSeek-V4.1-Flash & 98.3 & 98.3 & 95.4 & 99.2 & 96.9 & +0.0 & 96.2 & 93.3 & 98.1 & 35.6 & 30.5 & 40.9\\
\midrule
\multicolumn{13}{@{}l}{\textit{Open-weight models}}\\
Qwen3.8-27B & 98.5 & 98.3 & 95.8 & 98.7 & 95.6 & +0.2 & 97.5 & 93.5 & 98.3 & 72.3 & 68.7 & 76.1\\
Qwen3.8-Flash-Next & 98.5 & 98.5 & 95.6 & 99.2 & 95.8 & +0.0 & 98.8 & 95.6 & 98.7 & 75.5 & 70.2 & 81.1\\
Gemma 4 31B IT & 98.8 & 98.5 & 95.8 & 98.5 & 96.2 & +0.4 & 96.5 & 92.3 & 98.7 & 89.0 & 87.3 & 90.7\\
Qwen3.5-122B-A10B & 99.0 & 99.6 & 96.5 & 99.8 & 97.5 & $-$0.6 & 99.4 & 95.8 & 98.7 & 91.6 & 88.7 & 94.6\\
\midrule
\multicolumn{13}{@{}l}{\textit{Finance models}}\\
Fin-R1 & 82.7 & 73.5 & 81.6 & 75.0 & 63.5 & +9.2 & 79.1 & 63.9 & 72.6 & 27.5 & 23.6 & 31.7\\
Fin-o1-8B & 87.3 & 85.4 & 85.8 & 87.7 & 82.3 & +1.9 & 84.1 & 77.5 & 91.7 & 47.4 & 41.1 & 54.1\\
ODA-Fin-RL-8B & 90.6 & 89.1 & 91.0 & 93.3 & 90.2 & +1.5 & 92.1 & 85.6 & 90.8 & 45.7 & 41.1 & 50.6\\
Plutus-8B-Instruct & 77.5 & 74.3 & 78.5 & 74.7 & 68.9 & +3.3 & 72.0 & 74.7 & 76.4 & 27.2 & 26.9 & 27.4\\
\midrule
\multicolumn{13}{@{}l}{\textit{Base models of the finance models}}\\
Qwen2.5-7B-Instruct & 88.3 & 82.7 & 86.0 & 83.3 & 69.5 & +5.6 & 83.1 & 70.4 & 83.3 & 30.5 & 30.9 & 30.1\\
Qwen3-8B & 94.2 & 93.1 & 90.8 & 91.0 & 89.6 & +1.2 & 90.4 & 84.6 & 94.8 & 68.7 & 62.9 & 74.9\\
Llama-Krikri-8B-Instruct & 84.1 & 80.2 & 82.5 & 78.3 & 74.3 & +3.8 & 77.5 & 76.0 & 83.7 & 27.3 & 29.5 & 25.1\\
\midrule
\multicolumn{13}{@{}l}{\textit{Baselines}}\\
Random & 23.8 & 24.4 & 25.5 & 24.2 & 24.4 & $-$0.6 & 25.1 & 25.3 & 29.0 & 24.7 & 23.3 & 26.3\\
Majority / select-all & 24.2 & 24.2 & 24.2 & 24.2 & 24.2 & +0.0 & 24.2 & 24.2 & 24.2 & 24.9 & 24.0 & 25.9\\
\bottomrule
\end{tabular}}
\caption{\textbf{Language and label ablations} (accuracy, \%). T2a: joint EN+FR+EL evidence, without English, or a single language, with English labels (C1); $\Delta_\text{EN}$ = joint $-$ no EN. C2: French or Greek official labels without concept IDs on the single-language item; C3: English, French and Greek labels on the joint item. T2c: odd-one-out accuracy, split by whether another excerpt shares the odd excerpt's language.}
\label{tab:language}
\end{table*}

\begin{table}[!t]
\centering\footnotesize\setlength{\tabcolsep}{3.5pt}
\resizebox{\ifdim\width>\columnwidth\columnwidth\else\width\fi}{!}{%
\begin{tabular}{@{}lcccccc@{}}
\toprule
 & \multicolumn{3}{c}{Joint EN+FR+EL} & \multicolumn{3}{c}{No EN}\\
\cmidrule(lr){2-4}\cmidrule(lr){5-7}
Model & EN F1 & FR F1 & EL F1 & FR F1 & EL F1 & Macro-F1\\
\midrule
\multicolumn{7}{@{}l}{\textit{API models}}\\
GPT-5.6 Luna & 85.0 & 88.3 & 83.9 & 88.6 & 84.3 & 89.0\\
Claude Haiku 4.5 & 85.3 & 85.4 & 83.6 & 86.2 & 81.1 & 85.2\\
Gemini 3.8 Flash & 89.8 & 90.3 & 85.3 & 90.7 & 86.2 & 89.4\\
DeepSeek-V4.1-Flash & 83.8 & 85.0 & 82.3 & 83.8 & 79.3 & 84.5\\
\midrule
\multicolumn{7}{@{}l}{\textit{Open-weight models}}\\
Qwen3.8-27B & 68.4 & 69.0 & 66.9 & 74.4 & 73.2 & 70.7\\
Qwen3.8-Flash-Next & 74.3 & 76.6 & 75.2 & 82.1 & 79.9 & 81.1\\
Gemma 4 31B IT & 84.6 & 86.2 & 80.4 & 87.9 & 80.8 & 88.0\\
Qwen3.5-122B-A10B & 87.7 & 89.1 & 85.8 & 90.4 & 85.9 & 91.1\\
\midrule
\multicolumn{7}{@{}l}{\textit{Finance models}}\\
Fin-R1 & 72.4 & 65.2 & 53.8 & 65.2 & 50.4 & 62.1\\
Fin-o1-8B & 75.7 & 72.3 & 64.1 & 71.1 & 63.5 & 73.0\\
ODA-Fin-RL-8B & 81.2 & 78.2 & 73.7 & 79.4 & 71.8 & 79.3\\
Plutus-8B-Instruct & 24.6 & 17.0 & 18.2 & 28.0 & 23.9 & 18.4\\
\midrule
\multicolumn{7}{@{}l}{\textit{Base models of the finance models}}\\
Qwen2.5-7B-Instruct & 66.6 & 52.6 & 39.0 & 55.9 & 47.0 & 54.3\\
Qwen3-8B & 80.4 & 77.2 & 73.0 & 76.0 & 71.3 & 78.2\\
Llama-Krikri-8B-Instruct & 49.2 & 35.1 & 29.8 & 48.5 & 38.9 & 50.4\\
\midrule
\multicolumn{7}{@{}l}{\textit{Baselines}}\\
Random & 26.0 & 27.3 & 28.6 & 27.4 & 27.9 & 28.2\\
Majority / select-all & 30.7 & 31.6 & 34.1 & 31.6 & 34.1 & 34.0\\
\bottomrule
\end{tabular}}
\caption{\textbf{T2b by language} (F1 of the selected numbers per language, \%), with and without the English excerpt.}
\label{tab:t2b-language}
\end{table}

\begin{table}[!t]
\centering\footnotesize\setlength{\tabcolsep}{3.5pt}
\resizebox{\ifdim\width>\columnwidth\columnwidth\else\width\fi}{!}{%
\begin{tabular}{@{}lccccc@{}}
\toprule
Model & Acc. & Odd EN & Odd FR & Odd EL & Invalid\\
\midrule
\multicolumn{6}{@{}l}{\textit{API models}}\\
GPT-5.6 Luna & 49.4 & 40.3 & 49.4 & 63.4 & 0.0\\
Claude Haiku 4.5 & 59.9 & 57.5 & 62.5 & 60.7 & 0.0\\
Gemini 3.8 Flash & 95.1 & 95.5 & 94.6 & 95.2 & 0.0\\
DeepSeek-V4.1-Flash & 35.6 & 38.5 & 34.5 & 32.4 & 8.2\\
\midrule
\multicolumn{6}{@{}l}{\textit{Open-weight models}}\\
Qwen3.8-27B & 72.3 & 71.9 & 73.2 & 71.7 & 27.0\\
Qwen3.8-Flash-Next & 75.5 & 74.2 & 73.8 & 79.3 & 23.4\\
Gemma 4 31B IT & 89.0 & 86.9 & 90.5 & 90.3 & 0.0\\
Qwen3.5-122B-A10B & 91.6 & 91.9 & 91.1 & 91.7 & 0.7\\
\midrule
\multicolumn{6}{@{}l}{\textit{Finance models}}\\
Fin-R1 & 27.5 & 27.6 & 24.4 & 31.0 & 0.0\\
Fin-o1-8B & 47.4 & 46.6 & 48.2 & 47.6 & 0.0\\
ODA-Fin-RL-8B & 45.7 & 42.5 & 48.2 & 47.6 & 43.3\\
Plutus-8B-Instruct & 27.2 & 18.1 & 29.2 & 38.6 & 0.0\\
\midrule
\multicolumn{6}{@{}l}{\textit{Base models of the finance models}}\\
Qwen2.5-7B-Instruct & 30.5 & 30.8 & 17.9 & 44.8 & 0.0\\
Qwen3-8B & 68.7 & 70.1 & 67.9 & 67.6 & 0.2\\
Llama-Krikri-8B-Instruct & 27.3 & 32.6 & 26.8 & 20.0 & 0.0\\
\midrule
\multicolumn{6}{@{}l}{\textit{Baselines}}\\
Random & 24.7 & 22.2 & 29.2 & 23.4 & 0.0\\
Majority / select-all & 24.9 & 23.1 & 25.0 & 27.6 & 0.0\\
\bottomrule
\end{tabular}}
\caption{\textbf{T2c by the language of the odd excerpt} (accuracy and invalid rate, \%; chance 25\%).}
\label{tab:t2c}
\end{table}

\noindent\textbf{Evidence languages.}
In single-language T2a, Qwen2.5-7B scores 69.5\% on Greek versus 86.0\% on English; Fin-R1 scores 63.5\% versus 81.6\%. Greek-oriented Llama-Krikri-8B reaches 74.3\% on Greek. Reference models score 94.8--98.3\% on English-only and Greek-only items, compared with 97.3--99.8\% on French-only items. These differences reflect both language and context. Table~\ref{tab:t2b-language} reports T2b differences across evidence-language conditions. These differences are not always small, even for the reference models. For T2c, accuracy when the odd excerpt's language is repeated versus unique is 93.1\% versus 97.3\% for Gemini, 88.7\% versus 94.6\% for Qwen3.5-122B, and 62.9\% versus 74.9\% for Qwen3-8B.

\begin{table*}[!t]
\centering\footnotesize\setlength{\tabcolsep}{3.5pt}
\resizebox{\ifdim\width>\textwidth\textwidth\else\width\fi}{!}{%
\begin{tabular}{@{}lcccccccccc@{}}
\toprule
 & \multicolumn{4}{c}{Monetary} & & & & \multicolumn{3}{c}{On identified facts}\\
\cmidrule(lr){2-5}\cmidrule(lr){9-11}
Model & I/Dr & I/Cr & D/Dr & D/Cr & Per sh. & Shares & None & Period & Balance & Joint F1\\
\midrule
\multicolumn{11}{@{}l}{\textit{API models}}\\
GPT-5.6 Luna & 93.6 & 95.4 & 66.4 & 78.2 & 97.9 & 91.3 & 97.2 & 99.4 & 81.1 & 80.8\\
Claude Haiku 4.5 & 89.7 & 91.7 & 58.8 & 78.8 & 97.1 & 90.6 & 94.6 & 98.4 & 80.6 & 79.0\\
Gemini 3.8 Flash & 97.5 & 98.8 & 85.1 & 93.2 & 98.3 & 99.8 & 95.3 & 99.7 & 94.5 & 93.3\\
DeepSeek-V4.1-Flash & 90.2 & 93.0 & 67.2 & 80.1 & 97.4 & 94.7 & 93.8 & 99.2 & 82.5 & 81.4\\
\midrule
\multicolumn{11}{@{}l}{\textit{Open-weight models}}\\
Qwen3.8-27B & 94.7 & 84.0 & 39.6 & 56.7 & 90.4 & 61.7 & 50.0 & 99.3 & 95.7 & 66.2\\
Qwen3.8-Flash-Next & 94.7 & 81.3 & 38.9 & 51.1 & 93.9 & 57.6 & 48.4 & 99.5 & 94.8 & 63.2\\
Gemma 4 31B IT & 94.1 & 96.1 & 70.3 & 83.0 & 97.4 & 93.5 & 99.2 & 99.4 & 84.7 & 84.6\\
Qwen3.5-122B-A10B & 95.9 & 94.8 & 68.7 & 82.8 & 97.6 & 90.0 & 83.1 & 99.4 & 88.9 & 84.4\\
\midrule
\multicolumn{11}{@{}l}{\textit{Finance models}}\\
Fin-R1 & 30.1 & 41.3 & 27.5 & 33.2 & 21.6 & 16.0 & 48.1 & 62.9 & 54.2 & 31.1\\
Fin-o1-8B & 45.2 & 48.8 & 40.5 & 48.5 & 81.8 & 37.6 & 54.8 & 76.5 & 68.9 & 46.4\\
ODA-Fin-RL-8B & 77.2 & 74.3 & 57.3 & 62.8 & 93.9 & 47.3 & 76.0 & 89.1 & 78.2 & 66.3\\
Plutus-8B-Instruct & 6.8 & 9.3 & 7.4 & 11.0 & 15.6 & 6.4 & 31.6 & 76.7 & 53.3 & 9.2\\
\midrule
\multicolumn{11}{@{}l}{\textit{Base models of the finance models}}\\
Qwen2.5-7B-Instruct & 34.8 & 40.3 & 36.6 & 40.0 & 54.5 & 20.4 & 49.9 & 77.0 & 60.3 & 37.6\\
Qwen3-8B & 77.8 & 76.0 & 54.7 & 68.1 & 92.8 & 44.3 & 81.0 & 90.1 & 79.0 & 68.3\\
Llama-Krikri-8B-Instruct & 28.0 & 39.3 & 25.1 & 39.7 & 70.2 & 15.6 & 43.4 & 69.7 & 60.3 & 33.6\\
\midrule
\multicolumn{11}{@{}l}{\textit{Baselines}}\\
Random & 7.5 & 8.0 & 9.5 & 10.3 & 2.7 & 1.0 & 22.4 & 50.0 & 16.5 & 6.5\\
Majority / select-all & 0.0 & 0.0 & 0.0 & 51.8 & 0.0 & 0.0 & 0.0 & 67.3 & 66.2 & 39.3\\
\bottomrule
\end{tabular}}
\caption{\textbf{T1 per class} (F1, \%; I/D = instant/duration, Dr/Cr = debit/credit) and attribute accuracy on identified facts; joint F1 requires the ID, type, period and balance to be correct.}
\label{tab:t1-detail}
\end{table*}

\begin{figure}[tbp]
\centering
\includegraphics[width=0.8\linewidth]{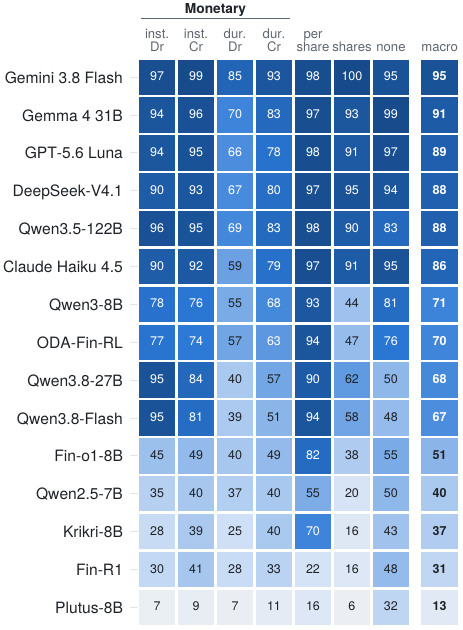}
\caption{\textbf{T1 by attribute class} (F1, \%). Monetary classes combine period (inst.\ instant, dur.\ duration) and balance (Dr debit, Cr credit); none: plainly non-financial numbers; macro: 7-class macro-F1. Models sorted by macro-F1.}
\label{fig:t1-classes}
\end{figure}

\noindent\textbf{Accounting attributes and format.}
Reference-model balance accuracy ranges from 80.6\% for Claude Haiku 4.5 to 95.7\% for Qwen3.8-27B. Debit-balance flow F1 is 85.1 for Gemini, 70.3 for Gemma, 68.7 for Qwen3.5-122B, 67.2 for DeepSeek, 66.4 for GPT-5.6 Luna and 58.8 for Claude Haiku 4.5; the class includes cash inflows, cash-flow reconciliation adjustments and distributions to owners. Instant-amount F1 is at least 89.7 for these models. Text versus table macro-F1 is 86.9 versus 95.4 for Gemini and 75.4 versus 90.6 for Gemma.

\begin{table*}[!t]
\centering\footnotesize\setlength{\tabcolsep}{3.5pt}
\resizebox{\ifdim\width>\textwidth\textwidth\else\width\fi}{!}{%
\begin{tabular}{@{}lccccccc@{}}
\toprule
 & T1 & T2a & T2b & T2c & \multicolumn{3}{c}{T1 by language}\\
\cmidrule(lr){6-8}
Specialist $-$ base model & F1 & Acc. & F1 & Acc. & EN & FR & EL\\
\midrule
Fin-R1 $-$ Qwen2.5-7B-Instruct & $-$8.4 & $-$5.6 & +9.9 & $-$3.0 & $-$12.1 & $-$5.8 & $-$7.0\\
Fin-o1-8B $-$ Qwen3-8B & $-$19.7 & $-$6.9 & $-$6.1 & $-$21.3 & $-$18.9 & $-$20.8 & $-$18.3\\
ODA-Fin-RL-8B $-$ Qwen3-8B & $-$0.9 & $-$3.6 & +0.6 & $-$23.0 & $-$2.1 & $-$0.1 & $-$0.5\\
Plutus-8B-Instruct $-$ Llama-Krikri-8B-Instruct & $-$24.7 & $-$6.5 & $-$34.2 & $-$0.2 & $-$27.6 & $-$19.6 & $-$25.4\\
\bottomrule
\end{tabular}}
\caption{\textbf{Finance specialization}: score of each finance model minus its base model (percentage points; T1 7-class macro-F1, T2a/T2c accuracy, T2b macro-F1).}
\label{tab:finance}
\end{table*}

\begin{figure}[tbp]
\centering
\includegraphics[width=0.8\linewidth]{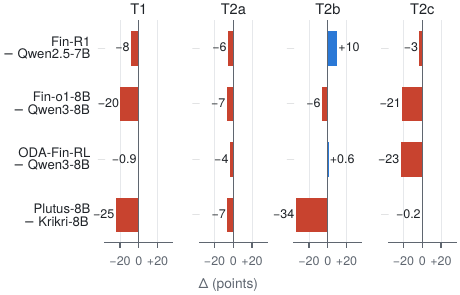}
\caption{\textbf{Finance models versus their base models} (points; T1 7-class macro-F1, T2a/T2c accuracy, T2b macro-F1). Red: the specialist scores lower; blue: higher. These are observed differences between released models under our protocol, not isolated effects of specialization. T1 by language in Table~\ref{tab:finance}.}
\label{fig:finance}
\end{figure}
\begin{table}[!t]
\centering\footnotesize\setlength{\tabcolsep}{3.5pt}
\resizebox{\ifdim\width>\columnwidth\columnwidth\else\width\fi}{!}{%
\begin{tabular}{@{}lcccccc@{}}
\toprule
 & \multicolumn{3}{c}{Joint EN+FR+EL} & \multicolumn{3}{c}{No EN}\\
\cmidrule(lr){2-4}\cmidrule(lr){5-7}
Model & 4 & 4 hard & 20 & 4 & 4 hard & 20\\
\midrule
\multicolumn{7}{@{}l}{\textit{API models}}\\
GPT-5.6 Luna & 97.3 & 94.8 & 89.4 & 97.3 & 95.6 & 89.8\\
Claude Haiku 4.5 & 97.5 & 95.6 & 90.6 & 98.3 & 96.2 & 91.2\\
Gemini 3.8 Flash & 99.4 & 98.7 & 97.5 & 99.2 & 98.8 & 97.5\\
DeepSeek-V4.1-Flash & 98.3 & 95.2 & 90.8 & 98.3 & 96.5 & 90.4\\
\midrule
\multicolumn{7}{@{}l}{\textit{Open-weight models}}\\
Qwen3.8-27B & 98.5 & 94.4 & 89.4 & 98.3 & 96.2 & 90.6\\
Qwen3.8-Flash-Next & 98.5 & 95.4 & 91.7 & 98.5 & 96.4 & 93.1\\
Gemma 4 31B IT & 98.8 & 97.1 & 93.3 & 98.5 & 95.8 & 93.9\\
Qwen3.5-122B-A10B & 99.0 & 97.9 & 96.4 & 99.6 & 98.7 & 95.4\\
\midrule
\multicolumn{7}{@{}l}{\textit{Finance models}}\\
Fin-R1 & 82.7 & 74.1 & 50.5 & 73.5 & 65.1 & 40.3\\
Fin-o1-8B & 87.3 & 87.7 & 73.3 & 85.4 & 84.5 & 67.2\\
ODA-Fin-RL-8B & 90.6 & 89.1 & 76.0 & 89.1 & 87.3 & 71.4\\
Plutus-8B-Instruct & 77.5 & 71.4 & 56.0 & 74.3 & 66.6 & 45.3\\
\midrule
\multicolumn{7}{@{}l}{\textit{Base models of the finance models}}\\
Qwen2.5-7B-Instruct & 88.3 & 83.9 & 65.8 & 82.7 & 75.2 & 52.8\\
Qwen3-8B & 94.2 & 93.1 & 84.1 & 93.1 & 91.9 & 81.6\\
Llama-Krikri-8B-Instruct & 84.1 & 78.5 & 63.7 & 80.2 & 75.4 & 53.0\\
\midrule
\multicolumn{7}{@{}l}{\textit{Baselines}}\\
Random & 23.8 & 23.0 & 5.6 & 24.4 & 24.6 & 5.6\\
Majority / select-all & 24.2 & 24.8 & 3.3 & 24.2 & 24.8 & 3.3\\
\midrule
Lexical overlap (no model) & 70.2 & 49.1 & 36.3 & \pending & \pending & \pending\\
\bottomrule
\end{tabular}}
\caption{\textbf{T2a with harder option sets} (accuracy, \%): the original four options, four harder options (calculation siblings and wording-matched distractors), and twenty options (chance 5\%), on the same items. Lexical overlap picks the option whose English label shares most words with the English rows of the marked numbers.}
\label{tab:t2a-options}
\end{table}

\noindent\textbf{Specialization and harder options.}
Relative to their bases, Fin-o1-8B and Plutus lose 19.7 and 24.7 points on T1. Fin-o1-8B and ODA-Fin-RL-8B lose 21.3 and 23.0 points on T2c, the latter largely reflecting budget exhaustion. All four specialists lose 3.6--6.9 points on T2a. Qwen2.5-7B scores 88.3, 83.9 and 65.8\% with the original four, harder four and twenty T2a options, respectively.

\subsection{Taxonomy-Label Comparisons}
\label{app:label-details}
C2 supplies official localized labels without concept identifiers, whereas C1 includes English labels and identifiers. On Greek-only items, switching from C1 to the Greek-label C2 condition lowers accuracy significantly for eight models: Qwen3-8B ($-$5.0 points, McNemar $p=0.002$), Fin-o1-8B ($-$4.8, $p=0.03$), ODA-Fin-RL-8B ($-$4.6, $p=0.002$), Gemma 4 31B ($-$3.8, $p<0.001$), DeepSeek-V4.1-Flash ($-$3.7, $p=0.003$), Claude Haiku 4.5 ($-$3.3, $p=0.006$), Qwen3.8-27B ($-$2.1, $p=0.03$) and Qwen3.5-122B ($-$1.7, $p=0.049$). Gemini 3.8 Flash and GPT-5.6 Luna change by $-$1.0 ($p=0.18$) and $-$0.8 ($p=0.56$), and only the Greek-adapted Plutus gains significantly ($+$5.8, $p=0.007$). French labels cause smaller losses (DeepSeek $-$3.1, $p<0.001$; Claude Haiku 4.5 $-$3.1, $p=0.001$; Gemma $-$1.9, $p=0.02$; GPT-5.6 Luna $-$1.5, $p=0.04$) and help Fin-R1 ($+$4.0, $p=0.03$).

Adding French and Greek labels to the English ones (C3) changes each reference-model score by at most 1.0 point, never significantly, and lowers the scores of Fin-R1 ($-$10.2) and Qwen2.5-7B ($-$5.0). Only Fin-o1-8B gains significantly ($+$4.4). These results compare the ways labels are presented to the model. C2 changes both label language and the availability of concept identifiers, so their effects cannot be separated.

\section{An Illustrative Grounding Item}
\label{app:example}
The following toy contexts are authored solely to explain the tasks and are not benchmark items. All amounts are in EUR millions.

\noindent\textbf{EN, issuer A.}
Revenue was [N1] 420; profit for the year was [N2] 41.

\noindent\textbf{FR, issuer B.}
Chiffre d'affaires : [N3] 280, contre [N4] 250 l'ann\'ee pr\'ec\'edente ; r\'esultat de l'exercice : [N5] 28.\\
\emph{Translation:} Revenue: 280, compared with 250 in the previous year; profit for the year: 28.

\noindent\textbf{EL, issuer C.}
{\greekfont Κύκλος εργασιών:} [N6] 130. {\greekfont Κέρδη χρήσης:} [N7] 12.\\
\emph{Translation:} Revenue: 130. Profit for the year: 12.

\noindent\textbf{T1.} Each amount is a monetary fact with period type \emph{duration}; revenue and profit have a \emph{credit} balance.

\noindent\textbf{T2a.} With N1, N3, N4 and N6 marked, the shared concept among (A) \concept{ifrs-full:Revenue}, (B) \concept{ifrs-full:ProfitLoss}, (C) \concept{ifrs-full:OtherIncome} and (D) \concept{ifrs-full:CostOfSales} is A.

\noindent\textbf{T2b.} Given \concept{ifrs-full:Revenue} with every fact numbered, the answer is \{N1, N3, N4, N6\}. Selecting \{N1, N3, N6\} misses the French comparative value; adding N2, N5 or N7 confuses revenue and profit.

\noindent\textbf{T2c.} Four excerpts mark revenue in three of them and profit for the year in the fourth; the model must name the fourth.
\section{Expert Review}
\label{app:review}
\noindent\textbf{Reviewers and procedure.} Two financial experts with \target{a professional accounting qualification (ACCA or CPA) and at least five years of IFRS reporting or audit experience} reviewed the sampled items. Their proficiency in French and Greek is \target{professional working level (CEFR C1) in both languages}. The web tool displayed benchmark excerpts alongside concept identifiers, English labels, IFRS definitions, and the taxonomy's presentation and calculation relations. Each tagged number was shown with its issuer-assigned concept. Experts saved a complete review for every sampled item independently, without seeing each other's decisions. They then discussed all disagreements and recorded joint decisions, which determine the reviewed subset's final annotations. Agreement is computed before discussion. The review took about \target{4} minutes per item.

\begin{table*}[!t]
\centering\footnotesize\setlength{\tabcolsep}{4pt}
\resizebox{\textwidth}{!}{%
\begin{tabular}{@{}lcccc@{}}
\toprule
 & T1 & T2a & T2b & T2c\\
\midrule
Items reviewed (10\% sample) & 131 & 52 & 52 & 53\\
Units judged & \target{4,213} numbers & 156 distractors & \target{2,108} facts & 212 excerpts\\
\midrule
Cohen's $\kappa$ (independent reviews) & \target{0.87} & \target{0.74} & \target{0.84} & \target{0.88}\\
Raw agreement (\%) & \target{98.6} & \target{88.5} & \target{97.3} & \target{94.3}\\
Items needing discussion (\%) & \target{22.1} & \target{19.2} & \target{25.0} & \target{9.4}\\
\midrule
Items with at least one correction (\%) & \target{14.5} & \target{28.8} & \target{15.4} & \target{5.7}\\
Units corrected (\%, 95\% CI) & \target{1.3 [0.8, 1.9]} & \target{12.8 [7.7, 18.6]} & \target{2.1 [1.1, 3.4]} & \target{3.8 [1.4, 6.6]}\\
Units judged ambiguous (\%) & \target{0.9} & \target{5.1} & \target{1.6} & \target{2.4}\\
Items dropped or invalid (\%) & \target{1.5} & \target{1.9} & \target{3.8} & \target{3.8}\\
\midrule
T2a distractors: too easy / good / hard but fair / ambiguous (\%) & & \multicolumn{3}{l}{\target{7.7 / 48.7 / 38.5 / 5.1}}\\
T2b difficulty: too easy / adequate / too hard (\%) & & & \multicolumn{2}{l}{\target{13.5 / 80.8 / 5.8}}\\
\midrule
Reference-model score change on audited items (points) & \target{0.6} & \target{1.2} & \target{1.4} & \target{0.9}\\
\bottomrule
\end{tabular}}
\caption{\textbf{Expert audit of a stratified 10\% sample per task.} Units are the judged elements: scored marked numbers (T1), distractors (T2a), numbered facts (T2b) and excerpts (T2c). Agreement is computed from the two independent reviews before discussion. Corrections and dropped items follow the joint final decisions. Score change: largest absolute difference, over the eight reference models, between scores on the audited subsets with original automatically derived and expert-corrected annotations (T1 7-class macro-F1, T2a and T2c accuracy, T2b macro-F1).}
\label{tab:review}
\end{table*}

\noindent\textbf{Sampling.} For each task, we randomly drew 10\% of the items within strata using a fixed seed. We stratified T1 by language and input format (table or text). For T2a/T2b, we stratified by data type and used three bands defined by terciles of the number of numbered facts. We sampled at most one group per concept. The same groups serve both tasks because they share contexts and targets. For T2c, the strata are the language of the odd excerpt and whether another excerpt shares that language. We report Wilson intervals for proportions computed over items. For proportions computed over facts, options or excerpts, we report 95\% percentile intervals from 2,000 bootstrap resamples of items. Resampling at the item level accounts for dependence among facts in the same item.

\noindent\textbf{Audit scope.} Each sampled item required checking every marked number, option or excerpt against IFRS definitions in three languages. The sample covers about 4,200 scored T1 numbers, 2,100 numbered T2b facts (400 targets), 156 T2a distractors and 212 T2c excerpts. The stratified audit estimates annotation error rates in the original automatically constructed benchmark.

\noindent\textbf{Guidelines.} Each task has separate guidelines, with the following rules in common. Reviewers judge the concept expressed by the visible evidence under the IFRS definitions, rather than accepting the issuer's choice. A number's sign, brackets, scale and currency do not change its concept. All columns of a row carry the same concept, including current and prior years, segments and equity components. When two concepts are defensible, reviewers choose \emph{ambiguous} and add a note. Ambiguous facts are excluded from scoring.

\textbf{T1 (numbers and attributes).} For every marked number, the expert answers: \emph{Does this number report a financial value that should be tagged, and are its type, period type and balance correct?} Labels: \emph{correct}; \emph{wrong attribute} (with the correct type, period type or balance); \emph{not a financial value} (for a gold fact, e.g.\ a year or reference that was tagged); \emph{missed fact} (for a number scored as \emph{none} that reports a financial value); \emph{ambiguous}. Period type follows the concept. Balances at a date are instants. Flows over the year, including movements in a balance, are durations. Balance follows the XBRL convention for the concept, regardless of the number's sign. Income, liabilities, equity and cash outflows are credit. Expenses, assets, cash inflows and distributions to owners are debit. Untagged amount-like numbers that are not scored are reviewed only if they are visibly identical to a tagged line (e.g.\ a repeated total). The item verdict is \emph{usable} or \emph{drop} (table cut so that rows or periods cannot be identified).

\textbf{T2a (concept and options).} The expert first checks whether every highlighted number reports the target concept. If not, the item is flagged and handled as in T2b. The expert then rates each of the three distractors against all groups of that concept. The ratings are \emph{too easy} (ruled out without reading the excerpts: different statement or nature, data type or direction, or an industry item that fits none of the issuers), \emph{good} (plausible at first sight, ruled out by the excerpts or definitions), \emph{hard but fair} (a sibling, current versus non-current, or a narrower or broader neighbour that is clearly wrong for every group), and \emph{ambiguous} (could also be correct for at least one group). Distractors rated too easy or ambiguous are replaced by a plausible but clearly wrong concept of the same data type, usually from the same statement. Option sets that give away the answer count as too easy. One example is a general concept presented alongside three industry-specific concepts. Test question: ``Would I accept this option if a model chose it?'' If yes or maybe, the distractor is ambiguous.

\textbf{T2b (labels of all facts).} For every numbered fact the expert answers: \emph{Is tagging this number with the target concept correct under IFRS?} Labels: \emph{is the concept}; \emph{not} (a different line item, a total that includes the concept, a component of it, a movement instead of a balance or vice versa, a different period type); \emph{ambiguous}. A number that equals a target in value but sits in another row is judged by its row label. When the issuer used a more specific concept that fits better, the expert labels that number \emph{not}. Each context receives a verdict: \emph{usable}, \emph{needs fix} or \emph{drop}. Reasons to drop a context include an incomplete table, missing headers or far too many numbers. A group is removed if any of its contexts is dropped. The expert also rates the difficulty of the group (too easy, adequate, too hard or too large).

\begin{table*}[!t]
\centering\scriptsize\setlength{\tabcolsep}{3.5pt}
\resizebox{\ifdim\width>\textwidth\textwidth\else\width\fi}{!}{%
\begin{tabular}{@{}llllll@{}}
\toprule
English issuer (filing country) & Sector & French issuer & Sector & Greek issuer & Sector\\
\midrule
KONE Corporation (FI) & Industrials & Safran & Industrials & Elastron & Materials\\
Airbus SE (NL) & Industrials & Thales & Industrials & Plastika Kritis & Materials\\
Bekaert (BE) & Industrials & Schneider Electric & Industrials & Alumil & Materials\\
STRABAG SE (AT) & Industrials & Legrand & Industrials & Flexopack & Materials\\
Heineken N.V. (NL) & Consumer & Bureau Veritas & Industrials & GEK TERNA & Industrials\\
Anheuser-Busch InBev (BE) & Consumer & Vinci & Industrials & AKTOR & Industrials\\
Kesko (FI) & Consumer & Bouygues$^\dagger$ & Industrials & AVAX & Industrials\\
Ferrari N.V. (NL) & Consumer & Eiffage$^\dagger$ & Industrials & Thessaloniki Port Authority & Industrials\\
ASML Holding N.V. (NL) & Tech./telecom & Air Liquide$^\dagger$ & Materials & Jumbo & Consumer\\
STMicroelectronics N.V. (NL) & Tech./telecom & Saint-Gobain & Materials & Fourlis & Consumer\\
Elisa (FI) & Tech./telecom & LVMH$^\dagger$ & Consumer & AS Company & Consumer\\
Koninklijke KPN N.V. (NL) & Tech./telecom & Hermès & Consumer & Loulis Food Ingredients & Consumer\\
Koninklijke Philips N.V. (NL) & Healthcare & Kering$^\dagger$ & Consumer & Quality and Reliability & Technology\\
Orion Corporation (FI) & Healthcare & L'Oréal & Consumer & Real Consulting & Technology\\
Recordati Industria Chimica e Farmaceutica S.p.A. (IT) & Healthcare & Danone$^\dagger$ & Consumer & Performance Technologies & Technology\\
ING Groep N.V. (NL) & Banking & Carrefour & Consumer & Space Hellas & Technology\\
KBC Group NV (BE) & Banking & Accor & Consumer & Athens Medical & Healthcare\\
BNP Paribas Group (FR) & Banking & Renault & Consumer & Lavipharm & Healthcare\\
NN Group N.V. (NL) & Insurance & Michelin & Consumer & Revoil & Energy\\
Sampo plc (FI) & Insurance & Dassault Systèmes & Tech./telecom & Elinoil & Energy\\
Shell plc (NL) & Energy/utilities & Capgemini$^\dagger$ & Tech./telecom & EYDAP & Utilities\\
Neste Oyj (FI) & Energy/utilities & Orange$^\dagger$ & Tech./telecom & EYATH & Utilities\\
VERBUND AG (AT) & Energy/utilities & Publicis$^\dagger$ & Tech./telecom & Alpha Trust Holdings & Financials\\
Italgas S.p.A. (IT) & Energy/utilities & Sanofi & Healthcare & Europe Holdings & Financials\\
UPM-Kymmene Corporation (FI) & Materials & Crédit Agricole & Banking & Dimand & Real estate\\
Akzo Nobel N.V. (NL) & Materials & Société Générale & Banking & Prodea Investments & Real estate\\
Outokumpu Corporation (FI) & Materials & AXA & Insurance & CPLP Shipping Holdings & Shipping\\
Unibail-Rodamco-Westfield SE (FR) & Real Estate & TotalEnergies & Energy/utilities & Safe Bulkers Participations & Shipping\\
Finnair Plc (FI) & Transport & Engie & Energy/utilities & Alter Ego Media & Media\\
PostNL N.V. (NL) & Transport & Veolia$^\dagger$ & Energy/utilities & Attica Publications & Media\\
\bottomrule
\end{tabular}}
\caption{\textbf{Issuers.} One FY2025 annual financial report (fiscal year ending 31 December 2025) per issuer. English: ESEF filings indexed by filings.xbrl.org; French: CAC~40 constituents, from filings.xbrl.org or, marked $^\dagger$, from the AMF's officially appointed mechanism; Greek: consolidated reports of Athens Exchange issuers. Sector labels are assigned manually.}
\label{tab:issuers}
\end{table*}

\textbf{T2c (odd excerpt).} The expert answers: \emph{Do the marked numbers of exactly three excerpts report the same concept, and those of the remaining excerpt a different one?} Labels: \emph{valid} (the three excerpts report $c$ and the odd excerpt reports $d$, which is clearly different under the definitions); \emph{ambiguous} ($d$ could also be reported as $c$ in that context, or one of the three excerpts could report $d$); \emph{invalid} (another excerpt is the odd one, or the marked numbers of an excerpt do not report its concept). The expert also checks whether each excerpt provides enough context, such as a row label or header, to identify its concept. If any excerpt lacks this context, the item is invalid.

\noindent\textbf{Results.} All main results use final annotations: expert corrections and exclusions for audited items, and automatically derived annotations elsewhere. Table~\ref{tab:review} summarizes the audit and compares scores using original and corrected annotations on reviewed items. These differences measure sensitivity to identified errors, not correctness of the whole benchmark.

\section{Collection and Preprocessing Details}
\label{app:construction}
\noindent\textbf{Report selection.}
We collected one FY2025 annual report (fiscal year ending 31 December 2025) from each of 30 issuers per language, with each issuer appearing in only one language (Appendix Table~\ref{tab:issuers}). \textbf{English} reports were selected from ESEF packages indexed as English by \texttt{filings.xbrl.org} and containing no validation errors. They cover ten sectors and six jurisdictions: the Netherlands 12, Finland 9, Belgium 3, Austria 2, Italy 2 and France 2. Three use US dollars. \textbf{French} reports cover French-reporting CAC~40 constituents as of 31 December 2025: 20 came from \texttt{filings.xbrl.org} and 10 from the AMF's officially appointed mechanism, \texttt{info-financiere.gouv.fr}. We excluded two issuers already in the English collection, one with a June year end, one without an FY2025 ESEF package, and foreign-registered constituents filing in English. \textbf{Greek} reports are consolidated filings from Athens Exchange issuers across eleven sectors, generally smaller companies than those in the other collections. We verified that every report covered a full year (a 365-day context ending on 31 December 2025) and assigned sectors manually.

\noindent\textbf{Evidence extraction.}
We render each inline XBRL report in a browser running without a graphical interface. HTML tables are converted to Markdown with their headers. On pages converted from PDF, we group text by position to recover lines. We then reconstruct paragraphs or tables by grouping numbers whose right edges align. Running text is split at sentence boundaries beyond 1,500 characters. Columns containing numbers but no tagged facts are dropped to exclude untagged parent-company figures alongside consolidated data. Normalization covers Unicode NFC, non-breaking and thin spaces, zero-width and control characters, soft hyphens, private-use characters, ligatures and the Unicode minus sign. Each tagged fact retains its visible-text offsets, concept, datatype, period type and balance. Attributes come from the IFRS Accounting Taxonomy 2024 or the issuer schema for extensions. Empty nil facts without visible text are excluded.

\noindent\textbf{T1 marking.} Tables exceeding 60 tagged facts are split by row with headers repeated. Every number-like string receives an identifier (\texttt{[N1] 1,234}); dash-only nil displays are not marked. A displayed number counts once even if it has multiple tags.

\noindent\textbf{T2a contexts and sampling.} A context is a passage or a table window containing at most five header rows and the three rows above and below the first target row. Contexts with another concept tagged on a target number are excluded. Groups are sampled with a fixed seed and without reusing an issuer for that concept within a language.

\noindent\textbf{T2a distractors.} Distractors share the target's datatype and an IFRS presentation section and have the closest English labels. We exclude calculation relatives within two steps, concepts tagged interchangeably on the same row label, and options with indistinguishable labels in any language. Options are fixed per concept, with balanced order.

\noindent\textbf{T2c distractors.} The distractor shares a parent in the calculation hierarchy or has a similar label. It must be neither an ancestor nor a descendant of the target, and issuers must not use the two interchangeably. Their labels must also differ in each language, as with lease payments and repayments of borrowings.

\section{Extended Related Work}
\label{app:related-work}
\noindent\textbf{Financial extraction and reasoning.}
FiNER and FNXL assign XBRL concepts to numbers in text. Instruction tuning with information about each concept extends this tagging approach to large label sets \citep{loukas-etal-2022-finer,sharma-etal-2023-financial,khatuya-etal-2024-parameter}. FinTagging separates extraction and typing from taxonomy linking across text and tables, and examines error propagation through this pipeline \citep{wang2026fintagging}. These resources test semantic precision and extraction recall within one language. Our paired evaluation examines the relationship between concept recognition and exhaustive numerical evidence recovery on matched multilingual contexts. We also test whether models distinguish related concepts without candidate labels. FinQA and TAT-QA test financial evidence use and arithmetic \citep{chen-etal-2021-finqa,zhu-etal-2021-tat}; FinQA also reports supporting-fact retrieval recall. We instead examine target coverage conditional on correct concept recognition, without requiring arithmetic.

\noindent\textbf{Multilingual evaluation.}
XLM-R, XTREME-R, and multilingual E5 address multilingual representations, transfer, and retrieval \citep{conneau-etal-2020-unsupervised,ruder-etal-2021-xtreme,wang2024e5}. MultiFin uses a shared set of financial topics across languages. MultiFinBen covers multilingual financial understanding, extraction and reasoning \citep{jorgensen-etal-2023-multifin,peng-etal-2026-multifinben}. We focus on individual numbers and the accounting concepts they express. We match excerpts from different issuers by concept, without requiring parallel text. This lets us relate concept recognition to target coverage on the same evidence and assess concept discrimination separately. Because these contexts come from different issuers, differences in coverage may also reflect sector and document difficulty, rather than language alone.

\end{document}